\documentclass[journal]{vgtc}                % final (journal style)
\ifpdf%                                % if we use pdflatex
  \pdfoutput=1\relax                   % create PDFs from pdfLaTeX
  \usepackage{graphicx}                % allow us to embed graphics files
  \DeclareGraphicsExtensions{.pdf,.png,.jpg,.jpeg} % for pdflatex we expect .pdf, .png, or .jpg files
\else%                                 % else we use pure latex
  \usepackage{graphicx}                % allow us to embed graphics files
  \DeclareGraphicsExtensions{.eps}     % for pure latex we expect eps files
\fi%

\graphicspath{{figures/}{pictures/}{images/}{./}} % where to search for the images

\usepackage{microtype}                 % use micro-typography (slightly more compact, better to read)
\PassOptionsToPackage{warn}{textcomp}  % to address font issues with \textrightarrow
\usepackage{textcomp}                  % use better special symbols
\usepackage{mathptmx}                  % use matching math font
\usepackage{times}                     % we use Times as the main font
\usepackage{cite}                      % needed to automatically sort the references
\usepackage{tabu}                      % only used for the table example
\usepackage{multirow}
\usepackage{booktabs}                  % only used for the table example
\DeclareMathAlphabet{\pazocal}{OMS}{zplm}{m}{n}

\onlineid{1186}

\vgtccategory{Research}
\vgtcpapertype{application/design study}

\newcommand{\sysname}{\textit{CNN}Pruner}

\title{\sysname{}: Pruning Convolutional Neural Networks\\ with Visual Analytics}

\author{Guan Li, Junpeng Wang, Han-Wei Shen, Kaixin Chen, Guihua Shan, and Zhonghua Lu}

\authorfooter{
\item
 Guan Li, Kaixin Chen, Guihua Shan, and Zhonghua Lu are with Computer Network Information Center, Chinese Academy of Sciences. They are also with University of Chinese Academy of Sciences.  E-mail:\{liguan, sgh, zhlu\}@sccas.cn, chenkaixin@cnic.cn.
\item
 Junpeng Wang is with Visa Research. E-mail: junpeng.wang.nk@gmail.com.
\item
 Han-Wei Shen is with The Ohio State University. E-mail: shen.94@osu.edu.
 \item 
 Guihua Shan is the corresponding author. 
}

\abstract{
Convolutional neural networks (CNNs) have demonstrated extraordinarily good performance in many computer vision tasks. The increasing size of CNN models, however, prevents them from being widely deployed to devices with limited computational resources, e.g., mobile/embedded devices. The emerging topic of \textit{model pruning} strives to address this problem by removing less important neurons and fine-tuning the pruned networks to minimize the accuracy loss. Nevertheless, existing automated pruning solutions often rely on a numerical threshold of the pruning criteria, lacking the flexibility to optimally balance the trade-off between efficiency and accuracy. Moreover, the complicated interplay between the stages of neuron pruning and model fine-tuning makes this process opaque, and therefore becomes difficult to optimize. 
In this paper, we address these challenges through a visual analytics approach, named \sysname{}. It considers the importance of convolutional filters through both \textit{instability} and \textit{sensitivity}, and allows users to interactively create pruning plans according to a desired goal on model size or accuracy. Also, \sysname{} integrates state-of-the-art filter visualization techniques to help users understand the roles that different filters played and refine their pruning plans. Through comprehensive case studies on CNNs with real-world sizes, we validate the effectiveness of \sysname{}.

}


\CCScatlist{ % not used in journal version
 \CCScat{K.6.1}{Management of Computing and Information Systems}%
{Project and People Management}{Life Cycle};
 \CCScat{K.7.m}{The Computing Profession}{Miscellaneous}{Ethics}
}

\teaser{
  \centering
  \includegraphics[width=\linewidth]{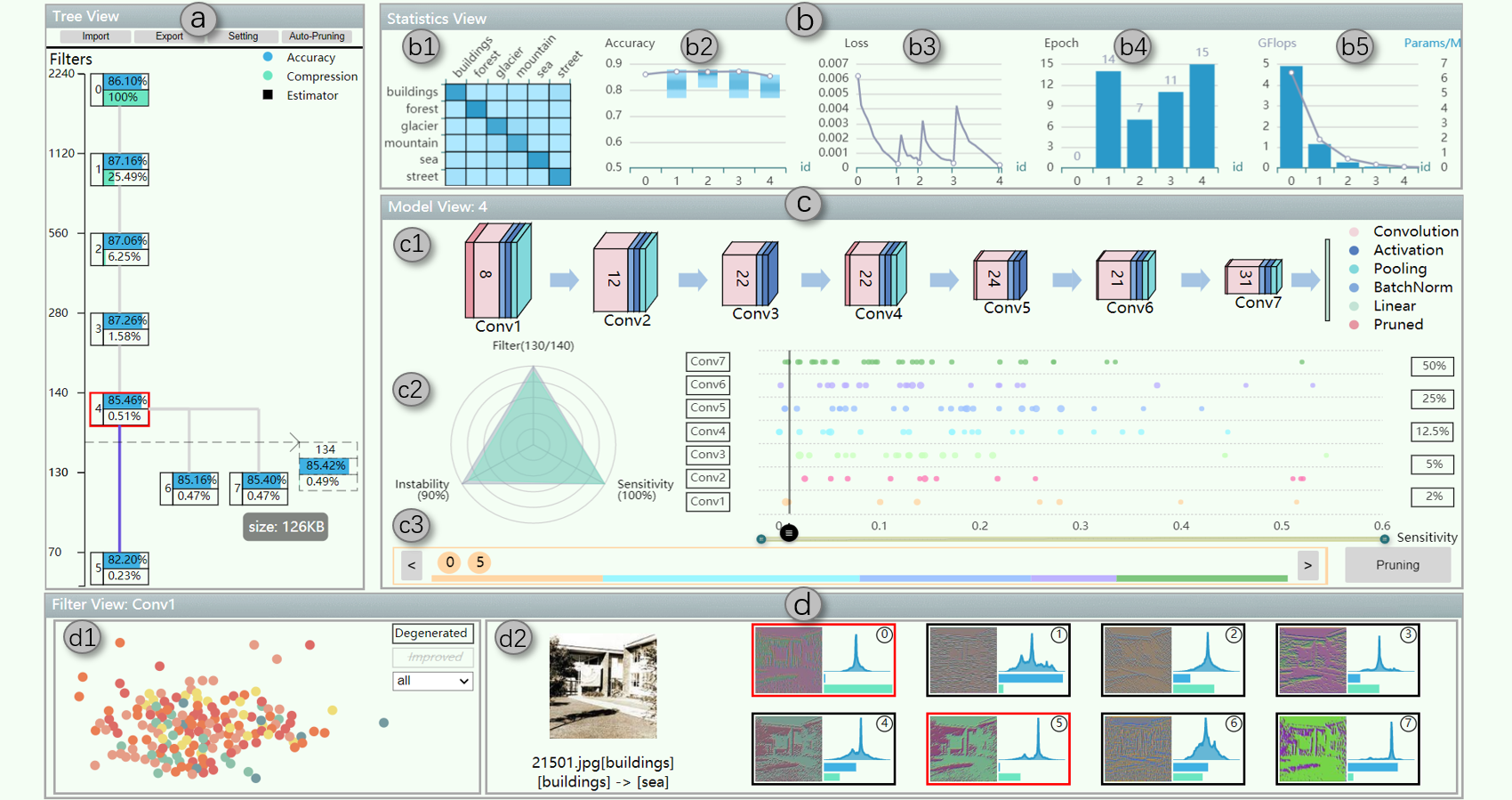}
  \vspace{-0.6cm}
  \caption{\sysname{}: (a) the Tree view helps to track different pruning plans; (b) the Statistics view presents model-critic statistics to monitor the pruned models; (c) the Model view enables users to interactively conduct the pruning with informative visual hints from different criteria; (d) the Filter view presents details of individual filters for users to investigate and interactively prune them.}
  \label{fig:teaser}
}

\vgtcinsertpkg

\begin{document}



\firstsection{Introduction}
\maketitle
% about cnns general intro of the problem
Convolutional neural networks (CNNs) have demonstrated extraordinarily good performance in many applications, such as image classiﬁcation, object detection, and speech recognition \cite{krizhevsky2017imagenet, simonyan2015very, girshick2014rich, lecun2015deep, seide2011conversational}. The recent improvements of CNNs' performance are often at the cost of model sizes. It becomes increasingly more common now to see models with hundreds of layers and millions of parameters. For example, VGG-16 \cite{simonyan2015very} is a commonly used model for classification tasks. It has ${\sim}138.3$ millions of parameters, and one forward pass of the model needs ${\sim}15.5$ billions of floating-point operations. The ever increasing size of CNNs, however, also prevents them from being widely deployed to devices with limited resources, e.g., mobile/embedded devices.

\begin{figure}
 \centering 
 \includegraphics[width=\columnwidth]{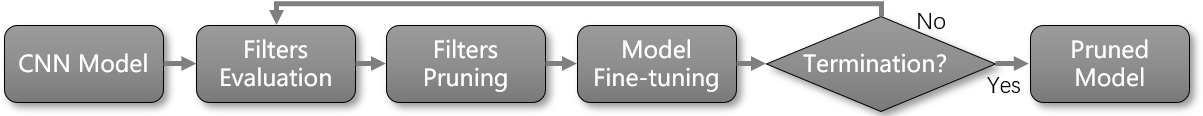}
 \vspace{-0.6cm}
 \caption{The iterative model pruning process of CNN models.} 
 \label{fig:prun_pipeline}
\end{figure}

% model pruning: specific problem focused on, core idea and standard process, schedule
Model pruning, i.e., compressing models' size by removing decision-irrelevant or less important parameters, aims to solve this problem. The idea can be traced back to the 1990s, when LeCun \textit{et al.} \cite{lecun1990optimal} first improved the efficiency of their neural networks by removing unimportant model parameters (weights) based on information theory metrics. In general, model pruning algorithms can be divided into structured pruning and unstructured pruning \cite{he2019filter}. Compared to unstructured pruning, which requires support from additional hardware to achieve excellent performance, structured pruning has gradually dominated the recent developments and has become a hot research topic. Most notably, filter pruning is an effective structured pruning method, which directly prunes filters that are less relevant to the prediction outcomes to reduce models' size. There exist three key steps in a typical filter pruning pipeline: 1) filters evaluation; 2) filters pruning; 3) model fine-tuning. Frequently, the pipeline is executed in an automated yet iterative manner (\autoref{fig:prun_pipeline}), where the filters are removed based on hard thresholds, and the models are pruned multiple times to achieve the desired compression goal, without significantly compromising the models' accuracy.

\setlength{\belowcaptionskip}{-12pt}
\setlength{\belowcaptionskip}{0pt}

% ----challenges, existing solution, few citations, limitation----
% ----the reviewer suggest rewriting this paragraph----
% ----the following text is the old one----

Nevertheless, existing automated CNN pruning solutions lack the flexibility to optimally balance the trade-off between pruning efficiency and prediction accuracy. The automated pruning usually removes a fixed number or a fixed percentage of convolutional filters in each pruning. If too many filters are removed in one pruning iteration, the model will be severely damaged and difficult to recover. Conversely, if too few filters are deleted, the effectiveness of pruning will be significantly undermined. In practice, the degree of ``over-parameterization'' is related to the corresponding CNN's size and the type of computer vision tasks, and in turn, different models have different abilities to recover from the damage caused by filter removal. Using a fixed numerical threshold as the criterion to remove filters in each pruning iteration ignores the characteristics of each model and may not lead to the optimal pruning solution. Moreover, the complicated interplay between the stages of filter pruning and model fine-tuning makes this process difficult to control, i.e., the automated pruning focuses solely on the accuracy of the pruned model, but pays little attention to the intermediate state changes. As a result, the anomalies accumulated in the iterative pruning process may get enlarged, which will affect the pruning efficiency and eventually impact the accuracy of the pruned model.

% contributions, in this work..
Focused on the above challenges, we propose \sysname, a visual analytics system to help deep learning experts create interactive pruning plans and evaluate the pruning process. \sysname{} contains four main visualization components (\autoref{fig:teaser}): (a) the {\em Tree View} helps to overview and track the altered models from iterative pruning stages; (b) the {\em Statistics View} presents the loss/accuracy fluctuation, model recovery capability, and recovery cost to help users adjust pruning strategy in time; (c) the {\em Model View}, facilitated with two new metrics of instability and sensitivity, evaluate the importance of different CNN filters and enables users to interactively create pruning plans; (d) the {\em Filter View} reveals the roles that different filters played in the prediction process and helps users interpret and prune the CNN model. We conducted case studies with \sysname{} on CNNs of real-world sizes to validate its effectiveness. To sum up, the contributions of our work are: 
\begin{itemize}
\itemsep-0.1em 
\item We design and develop a visual analytics system to help deep learning experts progressively analyze the CNN pruning process and introduce interactive intervention to the process on-demand.

\item We introduce two metrics (instability and sensitivity) to assist model designers in better estimating filters' importance before pruning, and three criteria (recovery capability, loss fluctuation, and recovery cost) to evaluate the pruned model.

\item We examine data instances, where the intermediate pruned models behave differently, to study the critical filters and interpret the internal working mechanism of CNNs. 
\end{itemize}

\section{Related works}
In this section, we introduce the concept of model pruning and its development in the field of deep learning. Also, we review the visual analytics works in interpreting and diagnosing deep learning models.

\subsection{Model Pruning}
Model pruning compresses deep learning models by removing less important parameters and seeks a trade-off between model size and prediction accuracy.
% The most critical step in model pruning is to remove the unimportant parameters while minimizing the accuracy difference between the original and the pruned model.
To date, many works have achieved good performance in neural network pruning, and we can roughly divide them into two categories~\cite{he2019filter}, weighted pruning and filter pruning.

% Weight pruning is an unstructured pruning method which deleted weights or compressed weights in a filter.
Weight pruning is an unstructured pruning method that deletes weights or compresses weights in a filter.
Han \textit{et al.} \cite{han2015learning} proposed a method to reduce the model size by removing the unimportant connections. They applied this method to CNN models that are trained for the ImageNet dataset, and reduced parameters about 89\% for the AlexNet model and about 92\% for the VGG-16 model.
Han \textit{et al.} \cite{han2016deep} used weight sharing on the basis of removing the unimportant connections \cite{han2015learning}, and employed the Huffman encoding to compress the weight to maximize the compression rate. Their experiment shows that through the compression of the VGG-16 model, the method can reduce the memory consumption from 552 MB to 11.3 MB without compromising the accuracy.
Carreira-Perpinan \textit{et al.} \cite{carreira-perpinan2018learning} proposed a method to find unimportant weights by minimizing loss changes while compressing those weights.
Some other studies have achieved good results through weight pruning \cite{guo2016dynamic, dong2017learning, tung2018clip, he2018soft, tai2016convolutional, zhang2016accelerating}, but weight pruning may cause unstructured sparsities and requires support from additional hardware to achieve excellent performance.

Filter pruning is a structured pruning method that directly removes convolutional filters from CNNs.
Luo \textit{et al.} \cite{luo2017thinet} proposed a framework named ThiNet to help the user identify the unimportant filters by computing the statistical information of adjacent layers.
Li \textit{et al.} \cite{li2017pruning} proposed an acceleration method for CNNs by removing the filters and their feature maps, which could reduce the computation to 34\% of the VGG-16 model.
Molchanov \textit{et al.} \cite{molchanov2017pruning} proposed a new Taylor expansion criterion to find the filters which have little influence on the loss value and remove them to reduce the model size.
Some other pruning studies along this line also show good pruning results \cite{liu2017learning, he2019filter, he2017channel, yu2018nisp, dubey2018coreset}.
Filter pruning keeps the regular structure of the model but significantly reduces the calculation and storage cost, making it a popular solution for model compression. We focus on this approach in this work as well.

Most of the aforementioned model pruning studies focus on proposing new pruning criteria and use a small fixed numerical threshold to determine the number of filters to be removed. This is because they mainly focus on the accuracy of the final model but concern less on the intermediate pruning process.
The small number of removal makes the model recover easily from the pruning and often leads to an optimal pruning result. However, it prolongs the pruning process, as more pruning iterations will be needed to achieve the compression goal. The process is usually not efficient and may incur a higher computational cost to perform fine-tuning in each pruning.

\subsection{Visual Analytics for Deep Neural Networks}
% Recently, it has been shown that visual analytics can significantly enhance our ability to perform analysis of complex models, especially deep neural networks (DNNs)~\cite{choo2018visual, yuan2021survey}.
% Visual analytics 
Based on the taxonomy from~\cite{choo2018visual, liu2017towards, yuan2021survey}, the visualizations for deep neural networks (DNNs) interpretations can roughly be categorized into three groups, targeting on model understanding \cite{liu2017towards, wang2018ganviz, kahng2019gan, ming2019rulematrix}, model debugging \cite{liu2018analyzing, pezzotti2018deepeyes, wang2019deepvid, zhang2019manifold, ren2017squares}, and model refinement \cite{bilal2018do, wang2018dqnviz}.

To understand a DNN model, researchers usually use visualization techniques to show the internal structure and state information of the model.
For example, CNNVis \cite{liu2017towards} uses directed acyclic graphs to formulate the model architecture and help domain experts understand CNN through visualization.
GANViz \cite{wang2018ganviz} helps the user understand the model by visualizing and comparing the internal model states (i.e., hidden activations) of the generative adversarial networks (GANs) \cite{goodfellow2014generative} over the training process.
GAN Lab \cite{kahng2019gan} is an interactive visualization tool for non-experts to learn the GAN models, and it significantly reduces the difficulty of understanding complex generative neural networks by using visualization techniques.

To debug/diagnose a DNN model, researchers usually define some visual evaluation methods to assist in the analysis of the model. For example, DeepEyes \cite{pezzotti2018deepeyes} helps the user diagnose a CNN model by visualizing the convolutional layers and convolutional filters. Based on the active level of different filters, this system improves the efficiency of model design by optimizing the network structure.
DGMTracker \cite{liu2018analyzing} monitors and diagnoses the training process of deep generative models through the visualization of a large amount of time-series information over time.

For model improvement, researchers usually use visualizations to help users identify the weakness of the model. For example, DQNViz \cite{wang2018dqnviz} exposes the details in the training process of deep Q-networks~\cite{mnih2015human} and uses visualization techniques to extract useful patterns of the model to better control the training.
Blocks \cite{bilal2018do} uses visualization techniques to analyze the impact of class hierarchy on the training of CNN models. Using the analysis results, the tool can accelerate model convergence and alleviate the problem of overfitting.

These studies have proved the effectiveness of visualization and visual analytics in the machine learning filed. Our work focuses on CNN model pruning and uses visualization to help deep learning experts to better understand and improve the pruning process of CNN models. We believe that with visualization and visual analytics, our system can effectively improve the efficiency of model pruning.

\section{Background And Concepts}
This section introduces the basic concepts of model pruning and a state-of-the-art filter visualization technique. Following them, we introduce the metrics used in this work and propose a novel evaluation concept.

\subsection{Pruning CNN Filters with Taylor Expansion}
This section describes the details of each step in the filter pruning process and introduces the Taylor expansion based filter evaluations.

\subsubsection{Taylor Expansion Criterion}
\label{sec:taylor}
Our work uses the Taylor expansion criterion~\cite{molchanov2017pruning} for filter pruning. Its idea is to remove filters and check how significant the removal will impact the loss function, i.e., examine the importance of filters by perturbation. The resulting importance values can then be used to prioritize filters during pruning. Mathematically, this process can be denoted as:
\begin{equation}
     {\Delta \pazocal{L}(f_i)} = \vert{\pazocal{L}(D, f_i = 0) - \pazocal{L}(D, f_i)}\vert
     \label{formula:cost}
\end{equation}
where $D$ is the training data, $\pazocal{L}()$ is the loss function, and $f_i$ is the output (i.e. feature map) produced from filter $i$, and $\pazocal{L}(D,f_i)$ is the loss before any model perturbation. $\pazocal{L}(D,f_i = 0)$ is the loss when $f_i$ is removed.

Physically removing individual filters and recomputing the loss for each removal is computationally expensive. But, the process can be approximated through Taylor expansions, as demonstrated in~\cite{molchanov2017pruning}, i.e.,
\begin{equation}
     {\pazocal{L}(D, f_i = 0)} = {\pazocal{L}(D, f_i) - \frac{\partial{\pazocal{L}}}{\partial{f_i}}{f_i}}
\label{formula:deltacost}
\end{equation}
${\Delta \pazocal{L}(f_i)}$ can then be transformed as follows:
\begin{equation}
     \Delta \pazocal{L}(f_i) = \vert{\pazocal{L}(D, f_i) - \frac{\partial{\pazocal{L}}}{\partial{f_i}}{f_i} - \pazocal{L}(D, f_i)}\vert = \vert{\frac{\partial{\pazocal{L}}}{\partial{f_i}}{f_i}}\vert
\label{formula:taylor}
\end{equation}

In \autoref{formula:taylor}, we need to calculate the product of the feature map and the gradient (the loss function w.r.t. to the feature map) to get the estimated cost of removing the corresponding filter, and this value can be calculated through back-propagation. After the calculation, $l2-$normalization is used to normalize the set of $\Delta \pazocal{L}$ values resulted from removing individual filters. With the normalized importance values, we can prioritize all filters and prune the less important ones.
We call this process of choosing a proper importance criterion to prioritize filters and decide the number of less important ones to remove as a \textbf{pruning plan}. Our objective is to derive efficient and effective pruning plans through interactive visual analytics.

\subsubsection{Model Fine-Tuning}
After removing the less important filters, the model structure is slightly damaged, and its accuracy will drop.
To recover the accuracy, we need to retrain the model using the training dataset.
As most of the important filters are still retained in the model, the original accuracy can be recovered with a few numbers of training epochs.
This process, i.e., retrain the CNN model to recover its accuracy, is called \textbf{fine-tuning}.

As described in \autoref{fig:prun_pipeline}, filters evaluation, filters pruning, and fine-tuning constitute \textbf{one pruning iteration}.
Repeating the process multiple times, we can generate the pruned CNN model.

\subsection{Filter Visualization Through Guided Back-Propagation}
Our primary goal in this work is to remove less important filters. Therefore, we need a proper filter visualization technique to reveal what individual features have been captured and to verify their importance. 
Guided back-propagation~\cite{springenberg2014striving}, as one of the state-of-the-art filter visualization technique, is adopted in our work.

Given an input image, this algorithm first performs a forward pass to the target network-layer.
It sets all activations of that layer to zero, except the one extracted by the filter that we want to analyze.
Next, the algorithm propagates the non-zero activations back to the input image to highlight what was extracted by the corresponding filter. Therefore, the resulting filter visualization image will have the same size as the input image and highlight what individual filters have captured.
We adopt this filter visualization technique, as it can work well in interpreting filters in deeper CNN layers \cite{springenberg2014striving}. It has also been adopted by multiple other model interpretation works~\cite{wang2018dqnviz}.

\begin{figure}[tbh]
 \centering 
 \includegraphics[width=\columnwidth]{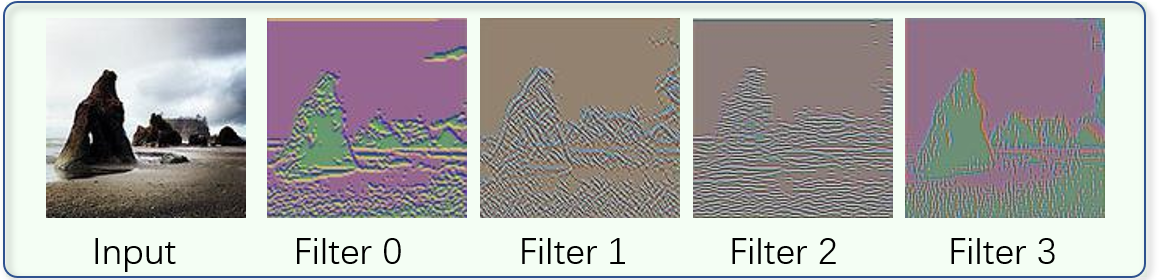}
 \vspace{-0.6cm}
 \caption{An example of filter visualization. The input is a mountain image. Filter 0 and Filter 3 capture the silhouette features of the mountain, whereas Filter 1 and Filter 2 capture the texture features of the mountain.}
 \label{fig:filer_demo}
\end{figure}

\autoref{fig:filer_demo} shows some filter visualization examples through this guided back-propagation technique. Four filters from Layer 0 of a 6-layer CNN is visualized, when taking a mountain image as input. From the highlighted regions in the filter visualization result, Filter 0 and Filter 3 capture the silhouette features of the mountain, whereas Filter 1 and Filter 2 capture the texture features of the mountain. 

\setlength{\belowcaptionskip}{-8pt}
\setlength{\belowcaptionskip}{0pt}

\subsection{Sensitivity and Instability}
Based upon the Taylor expansion algorithm explained in \autoref{sec:taylor}, we define one criterion and propose a new metric as another criterion for filter pruning in our work.

The \textbf{Sensitivity} of a filter reflects the filter's impact on the model's loss when being removed. It is calculated using L2-normalized $\Delta \pazocal{L}$ (Equation \ref{formula:taylor}). A filter with a lower ${Sensitivity}$ value should be removed first to reduce the impact on the model. 

Notice that repeating the sensitivity calculation for the same filter multiple times may result in different sensitivity values of the filter, due to the randomness inherited from the statistical parameter update process.
Specifically, the updates of CNN model parameters are often in the unit of data batches. Feeding data batches into a CNN with shuffled orders will result in different parameter update orders and scales.
The impact of this randomness is usually marginal to important filters, as their sensitivity values are always large. 
However, for less important ones, their sensitivity values are minimal and can be easily influenced by this randomness. Therefore, for these less important ones, their sensitivity orders may be very different from different calculations.

We introduce the metric \textbf{Instability} to accommodate the above issue, which is defined as the mean absolute deviation of the filters' ranks from different calculations, i.e.,
\begin{equation}
     Instability(f_j) = \frac{\sum_{i=0}^{n}{\vert(Rank_{i}(f_j) - \overline{Rank(f_j)})\vert}}{n}
\end{equation}
where $n$ is the total time that we computed the sensitivity for individual filters, $Rank_{i}(f_j)$ is the ranking of the $j$th filter in the $i$th computation, and $\overline{Rank(f_j)}$ is the average rankings for filter $j$.
The instability of a filter reflects the uncertainty of the removal order, and often, the filter with a higher instability indicates it is less important.
We set $n{=}5$ in this work, but it can be adjusted on-demand.

% instability is an interesting metric to capture how much the ranking fluctuates due to shuffled data, but why does it indicate if a filter is important? The authors need to substantiate the claim.

\subsection{Degenerated Instances and Improved Instances}

Each pruning iteration improves and degenerates the model a little bit, and its prediction accuracy also changes, i.e., some data instances in the test data have different predictions results from the original and the pruned model. To better index the subset of instances with different predictions from the two models, we define the following two concepts:

\noindent\textbf{Degenerated Instances} are images that are correctly predicted in the original model but incorrectly predicted by the pruned model, i.e., the pruning hurts the model's recognition ability on these images.

\noindent\textbf{Improved Instances} are images that are incorrectly predicted by the original model but correctly predicted by the pruned model, i.e., the pruning improves the model's recognition ability on these images.

The test dataset used by a CNN model usually contains many images, and it is difficult to analyze the effect of the pruning on every single image. The degenerated instances and improved instances help users to quickly locate the analysis target from the massive images, which improves the analysis efficiency.

% The reviewer sugguest delete this figure[first round][2020-07-18]
%\begin{figure}[tbh]
% \centering 
% \includegraphics[width=\columnwidth]{figures/degeneration_evolution.png}
% \caption{We were pruning Model A to get Model B. The blue circle indicates the images that can be correctly classified by Model A, and the green circle indicates the images that can be correctly classified by Model B.}
% \label{fig:degeneration_evolution}
%\end{figure}

\section{Design Requirement}
%We worked closely with a group of deep learning researchers and identified the following requirements. 

% \clr{We worked with a group of deep learning researchers and have some discussions during the system design stages. In addition, we also investigate some previous works about model pruning. Then, we identified the following requirements.}

We worked with a couple of deep learning researchers and had some discussions/interviews with them during the system design stages. Also, we investigated the related works on model pruning to identify the challenges that deep learning experts are facing with. 
From these discussions and literature reviews, we found that proposing effective pruning criteria is an important research topic, and the criteria are often evaluated by the accuracy of the pruned model. Based on different criteria, people often use a fixed number or a mathematical formula to decide the amount of filters to be removed in each pruning iteration, which lacks flexibility and is usually not efficient. For example, a small removal count is often used to guarantee the model's recovery capability. However, the small number often leads to more pruning iterations, which inevitably prolongs the pruning process, costing more computing resources for model fine-tuning. Additionally, we noticed that even if the original and pruned models have similar prediction accuracy, their recognition power for different classes may be very different. Revealing these details, along with other model-level details (e.g., model architecture evolution, recovery capabilities from pruning) are very important to understand the pruning process. 
Through the responses from the experts and our studies of the existing works, we have identified the following design requirements for \sysname{}.

\begin{itemize}
\setlength{\itemsep}{0pt}
\setlength{\parsep}{0pt}
\setlength{\parskip}{0pt}

\item \textbf{R1: Display different levels of information about the CNN models during pruning.} Many intermediate CNN models are generated in the iterative process of model pruning, and our system needs to track and display the details of those models. Displaying these model information is the basis for understanding and exploring the pruning process, which requires {\sysname} to: 

\begin{itemize}
\setlength{\itemsep}{0pt}
\setlength{\parsep}{0pt}
\setlength{\parskip}{0pt}
\item R1.1: track the intermediate models generated over the pruning process and index the models effectively.

\item R1.2: display the states of the pruned models and monitor the evolution of these states over the pruning process.

\item R1.3: visualize the internal structure of a selected CNN model (e.g., the original/intermediate/final pruned model) and its filters' attributes.

\end{itemize}

\item \textbf{R2: Interactively analyze and decide the number of filters to be removed in each pruning iteration.} After each pruning, the model needs to be fine-tuned, and its prediction accuracy will change. The experts want to minimize the computational cost for fine-tuning but restore the accuracy as much as possible. Therefore, they expect {\sysname} to help them analyze the impact of pruning and select the appropriate removal amount in each pruning. We, therefore, design {\sysname} to be able to: 

\begin{itemize}
\setlength{\itemsep}{0pt}
\setlength{\parsep}{0pt}
\setlength{\parskip}{0pt}
\item R2.1: estimate the influence of a pruning plan on the model before the pruning actually happens (i.e., pre-estimation).

\item R2.2: evaluate the quality of the pruning plan and the pruned model after each pruning (i.e., post-evaluation).

\item R2.3: assist the user in better selecting or optimizing the number of filters to be removed in each pruning iteration.

\end{itemize}

\item \textbf{R3: Understand model pruning process and refine the pruning plan.} The convolutional filters are the basic units to be removed in each pruning. The in-depth analysis of them can help the user better understand the pruning process and identify the abnormal changes of the accuracy values for different classes of the studied dataset. Therefore, {\sysname} needs to be able to:

\begin{itemize}
\setlength{\itemsep}{0pt}
\setlength{\parsep}{0pt}
\setlength{\parskip}{0pt}
    
\item R3.1: visualize the filters of interest and help the user understand the roles that different filters played during pruning.

\item R3.2: interactively refine the pruning plan by adding or removing filters to be pruned to reduce undesired changes of the model over the pruning.

\end{itemize}
\end{itemize}

\section{System Overview}
\begin{figure}[tbh]
 \centering 
 \includegraphics[width=\columnwidth]{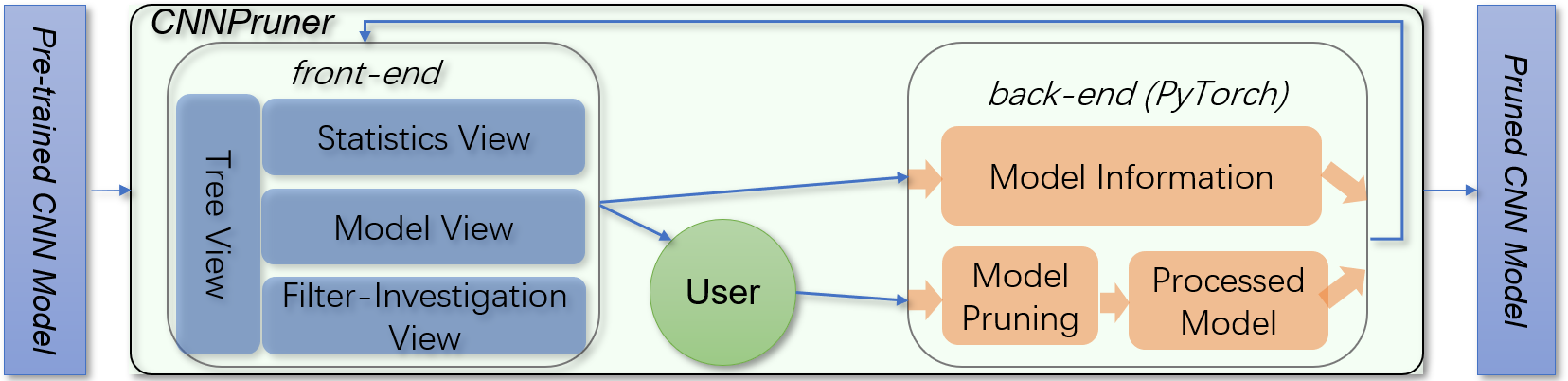}
 \vspace{-0.6cm}
 \caption{The architecture of {\sysname}, including a back-end powered by PyTorch and a web-based font-end visualization interface.}
 \label{fig:pipeline}
\end{figure}

\autoref{fig:pipeline} shows the architecture of {\sysname}, which contains a back-end powered by PyTorch \cite{PyTorch_framework}, and a web-based front-end for visualization and interaction.
We use the Flask \cite{flask_web} library to support the communication between the back-end and the front-end.

\sysname{} takes a pre-trained CNN model as input and outputs the pruned model. Users can flexibly interact with the four visualization components from the font-end to complete the above process.
In detail, the \textit{Tree} view layouts the pre-trained (tree root) and post-pruned CNN (tree leaf), as well as all intermediate pruned models through a tree structure (R1.1). An estimator (R2.3) is equipped in this view to help users estimate a proper number of filters to be removed between adjacent tree nodes (i.e., CNN models).
The \textit{Statistics} view (\autoref{fig:teaser}-b) shows the evolution of the model's statistics over the process of pruning (R1.2), where users can evaluate the pruning scheme through these statistics (R2.2).
The \textit{Model} view (\autoref{fig:teaser}c) presents the internal structure and the filter attributes of a selected tree node (i.e., a CNN model) from the \textit{Tree} view (R1.3).
It is the main component that allows users to interactively prune the selected model and provides immediate feedback to the pruning operation to guide users towards an optimal pruning plan.
The \textit{Filter} view (\autoref{fig:teaser}d) presents details of the individual filters to help users interpret them and interactively refine the pruning plan (R3.1, R3.2).
All of the four visualization components are coordinated, and they work together to meet the objective of helping experts understand, diagnose, and refine the pruning process of CNNs.

\section{Visual Analytics System: \sysname}
%four visualization components:
%   for each one : goal, visual encoding, interaction

\sysname{} (\autoref{fig:teaser}) is composed of four visualization components, demonstrating different levels of CNN information and the pruning process.
We provide the details of individual components in this section.

\subsection{\textit{Tree} View}
The \textit{Tree} view provides an overview of the iterative model pruning process (R1.1). The root and leaves of the tree are the original and the pruned models respectively. Each branch of the tree (connecting the root to a leaf) chains a sequence of intermediate models from the iterative pruning process. 
For each node, we use two horizontal filled rectangles to denote the corresponding model's prediction accuracy and compression ratio, and one vertical rectangle to display the model ID. The system automatically generates the ID, and the ID of the root model is $0$.
The vertical position of a node is decided by the number of filters in the corresponding model (see the left vertical axis). The edges connecting a pair of parent-child nodes represent the fine-tuning process, where we use gray and purple lines to indicate if the fine-tuning process converged or not after reaching the user-specified stop conditions (e.g., the maximum number of epochs). The \textit{Tree} view can track the whole pruning process, and each node in the tree represents a model. When the mouse hovers over a node, a prompt bar will appear to show the storage size of the corresponding model. The user can click on individual nodes of the tree to update the data displayed in other views. The node with the red border is the currently selected one.

The \textit{Tree} view is also equipped with a pruning estimator, to help users balance the trade-off between model size and prediction accuracy (R2.3). It estimates the number of filters in a model and the model's prediction accuracy, by linearly interpreting these values from a pair of most adjacent nodes from the tree. This rough linear estimation works well based on our experiments, and we expect more sophisticated interpolation algorithms to yield better results. The pruning estimator node is only visible when users pressing the black box icon on the top right corner of the tree view, and users can flexibly drag it vertically to generate the estimations dynamically.

% usage 
This view has two important parameters to be configured before any pruning (using the buttons on the top of this view). One is the pruning mode, which could be automated or manual pruning. The other is the termination criteria for fine-tuning. They are explained as follows:

\noindent\textbf{Auto/Manual-Pruning.} For auto-pruning, users specify a fixed ratio of the filters to be removed, e.g., 1/2, 1/3, or 1/4 of the total amount, and \sysname{} will iteratively remove the specified amount of filters (based on the pruning criteria) and fine-tune the model. The iterative process runs until the pruned model fails to meet the desired need (e.g., the prediction accuracy no longer meets the requirement). This process is automated but lack of pruning flexibility. Conversely, in manual-pruning, users can flexibly specify the number of filters to be removed (based on their distribution) in individual pruning iterations. 

\noindent\textbf{Termination Criteria for Fine-Tuning.} \sysname{} has three termination criteria to finish the fine-tuning process, (1) Delta Loss, i.e., the change of loss values, (2) the Target Accuracy, and (3) the Maximum Epoch. The fine-tuning will be terminated if any of them is met.

\subsection{\textit{Statistics} View}
The \textit{Statistics} view (Fig. \ref{fig:teaser}-b) is used to display detailed statistical information of the CNNs (R1.2).
When the user selects a model in the \textit{Tree} view, the system will find a path from the current node to the root node where the models along the path form a pruning process. 
This view shows statistical information from multiple dimensions over the pruning process.
There are five components in this view. 

\textbf{Confusion Matrix.}
\autoref{fig:teaser}-b1 shows the confusion matrix of the current model (R1.2). 
Diagonal cells of the matrix represent the accuracy of true-positive instances (i.e., the percentage of correctly predicted images in one category). 
Non-diagonal cells represent the percentage of incorrectly predicted images.
Values (of the cells) from small to large are mapping to colors from light blue to dark blue.
Clicking any cell of the matrix will show a line-chart presenting the value changes for the corresponding cell during the pruning process (i.e., X-axis is the pruning iterations, Y-axis is the cell values from different iterations). The line-chart reflects the model's prediction power for a particular category across the pruning process.

\textbf{Recovery Capability.} 
This sub-view (\autoref{fig:teaser}-b2) reveals the model's recovery capability after each pruning (R1.2), i.e., how difficult it is to learn the prediction power back over the fine-tuning process.
The X-axis is the model ID, and it represents different pruning iterations, whereas the Y-axis denotes the model's prediction accuracy from individual iterations.
The gray-curve connects the model's final accuracy values after the individual fine-tuning process. The rectangular color stripe at each iteration shows the distribution of the accuracy values from different epochs of the corresponding fine-tuning. A longer stripe indicates a more significant accuracy change before and after the fine-tuning. 
If the pruned filters have little effect on the model, the recovery region will be very short, and the accuracy fluctuation will be very small (i.e., a short strip with dark blue color).
The information from this sub-view is an important criterion to evaluate the pruning plan.

\textbf{Loss Fluctuation.} 
The \textit{Loss Fluctuation} sub-view shows the loss changes in the process of fine-tuning (R1.2).
The X-axis in the chart is the model ID, and the Y-axis is the loss value.
The curve between the two IDs represents the fluctuation of the training loss in the fine-tuning process (between two models).
The importance of a filter is estimated based on how significantly the loss will change when removing it.
The loss values, quantifying the inconsistency between the predicted label and the true label, can effectively monitor the model's evolution.
If our pruning plan is good enough, the impact on the loss will be small.
Therefore, the fluctuation of loss is another important criterion to measure the pruning plan, and this sub-view helps users analyze the fine-tuning process by displaying the loss fluctuation.

\textbf{Recovery Cost.} 
The \textit{Recovery Cost} sub-view shows the number of epochs in the fine-tuning process through a bar chart (R1.2).
The X-axis of the chart is the model ID, and the Y-axis is the epoch count.
If the pruning plan has little effect on the model, then only a small amount of training epochs is needed in the fine-tuning process to recover the accuracy.
Conversely, if over-pruning happens, even with a lot of training epochs, it is still difficult to recover the accuracy.
Therefore, the recovery cost is also a criterion to evaluate the pruning plan.
Through this sub-view, the user can have an intuitive understanding of the recovery cost in the pruning process.

\textbf{Parameters and Computation.} 
This sub-view displays the reduced amount of the model parameters and the computational cost. 
As is shown in Fig. \ref{fig:teaser}-b5, the line chart displays the reduced number of parameters, and the histogram displays a reduced amount of computations (R1.2).
The pruning process removes filters from the network, thus reducing the size of parameters.
Meanwhile, the size of the parameters is proportional to the amount of computation in the model.
By calculating the amount of computation for the process of one image in the test dataset, users can estimate the running efficiency of the model in mobile/embedded devices.
The user can verify if the pruned model meets the computation requirements or not through this chart.

All sub-views, except the Confusion Matrix, can be scaled horizontally to take the full space of the Statistics view (by double-clicking the corresponding sub-view). 
This interaction helps to scale the system when the pruning process is long or involves many pruning iterations.
It also reduces the information that users need to watch at once, helping them to better focus on a single metric at a time (rather than overwhelmed by all five statistical metrics).

\subsection{\textit{Model} View}
The visualization of the internal information of a CNN model can help users understand the state of CNN and make proper pruning plans. As shown in Fig. \ref{fig:teaser}-c, we designed the \textit{Model} view to display the architecture of the studied CNN model (\autoref{fig:teaser}-c1), the evaluation of filters from the model (\autoref{fig:teaser}-c2), and the pruning plan (\autoref{fig:teaser}-c3).

The architecture of the model selected in the \textit{Tree} view is displayed in Fig. \ref{fig:teaser}-c1 (R1.3, R2.1).
Each box in the architecture diagram represents a layer of the model.
Different colors represent different types of layers.
In particular, we use the red box to represent the deleted filter, and the width of this box to denote the percentage of the deleted filters in the current convolutional layer.
The height of each box is proportional to the size of the feature map.
The number on the box is the number of filters in the corresponding convolutional layer.

The visualization of filter evaluation is shown in Fig. \ref{fig:teaser}-c2, which consists of a radar plot and a bubble plot (R1.3, R2.1).
The radar plot shows the impact of the pruning plan on the current model.
There are three dimensions of information in the radar chart, namely, the number of filters, the remaining sensitivity percentage, and the remaining instability percentage.
The remaining percentage means the ratio of the metric between the model after this pruning iteration and the current model.
The bubble plot on the right shows the sensitivity and instability of each filter in each layer.
In the bubble plot, each bubble represents a filter, and different layers have different colors.
The X-axis represents the sensitivity value, and hence the bubbles closer to the right are the filters with more impact on the loss (i.e., important ones that should not be pruned).
The size of the bubbles represents the corresponding filters' instability, i.e., bigger ones correspond to larger values.

The pruning plan is shown in Fig. \ref{fig:teaser}-c3, and it shows the indices of filters that each layer will be removed (R2.1).
Each circle represents one filter, and the number on the circle is the index of the filter.
The circles of different layers use different colors, which are consistent with the bubble plot above.
The multi-color line under the circles is an overview of the number of filters to be removed in the pruning plan.
Different colors represent different convolutional layers, and the length of the color segment represents the percentage of the removed filters in the corresponding convolutional layer.

% interactive
This view displays information of the model selected from the \textit{Tree} view.
The icons (i.e., the layer legends) on the right of the model architecture support the filtering of different layers. For example, when clicking the icon for convolutional layers, other layers, e.g., pooling and linear layers, will become transparent to help users better focus on the layers in the analysis.
There is a vertical slider in the bubble plot, and users can drag it to specify the pruning threshold.
The bubbles on the left of the slider are shown in the pruning plan and represent the filters that will be removed in the current pruning.
Meanwhile, the radar plot on the left shows the influence of pruning on the number of filters, the sensitivity, and the instability (R2.1).
Dragging the slider will also change the width of the red boxes in the model architecture diagram and the proportion of different colors in the multi-color segment of the pruning plan (R2.1).
Additionally, the system provides a set of buttons on the right of the bubble plot to help users quickly move the slider to certain positions.
Users can scale the bubble plot horizontally along the sensitivity axis to reduce the occlusion between bubbles.
They can also switch among different convolutional layers in the \textit{Filter} view through the convolutional buttons between the radar plot and the bubble plot.

\subsection{\textit{Filter} View}
The \textit{Filter} view allows the user to conduct an in-depth analysis of a specific convolutional layer (R3.1, R3.2).
As shown in \autoref{fig:teaser}-d, this view consists of a scatter plot and a filter visualization matrix.
The points in the scatter plot represent the degenerated and the improved instances in the test dataset, and the color represents the category of the exemplars.
We use the t-SNE \cite{maaten2008visualizing} algorithm to process the image instances, and display them in the scatter plot.
Our system uses the degenerated and improved instances to distinguish sensitive images, which efficiently narrows down the analysis scope.
The selected image in the middle of the \textit{Filter} view shows the point that the user clicked in the scatter plot.
There are two lines of texts at the bottom of the image. The first line shows the image name and its true label.
The second line shows the labels of the image before and after the pruning, separated by an arrow.
In the filter visualization matrix, each item represents a filter, and the items with red borders will be deleted in current pruning.
The image in each item is the visualization of the filter.
The area chart on the top right of the item shows the distribution of pixel values of the filter visualization images.
The blue and green bar below the area chart represent the sensitivity and instability of the filter, respectively.

% interactive
When the user selects a node in the \textit{Tree} view, the system retrieves the degenerated and improved data instances according to the selected node and its child node.
The user can switch the displayed convolutional layer in the \textit{Layer} view by clicking on the convolutional buttons in the \textit{Model} view (between the radar plot and the bubble plot).
The scatter plot supports the filtering of different types of data instances through the icons on the upper right corner.
After the user clicks one point in the scatter plot, the selected image and the matrix view on the right will be updated accordingly to reflect the selection.
From the matrix view, the user can double-click any item to add/delete the corresponding filter to/from the current pruning plan.

\section{Case Studies}
In this section, through three cases we present how \sysname{} can assist pruning, improve pruning efficiency, and optimize pruning plans.
%%%%%%%%%%%%%%%%%%%%%%%%%%%%%%%%%%%%%%%%%%%%%%%%%%%%%%%%%%%%%%%%%%%%%%%%%%%%%%%%%%%%%%%%%%%%%%%%%%%%%

\subsection{Automated Pruning: A Guiding Example with MNIST}

\begin{figure}[t]
 \centering 
 \includegraphics[width=\columnwidth]{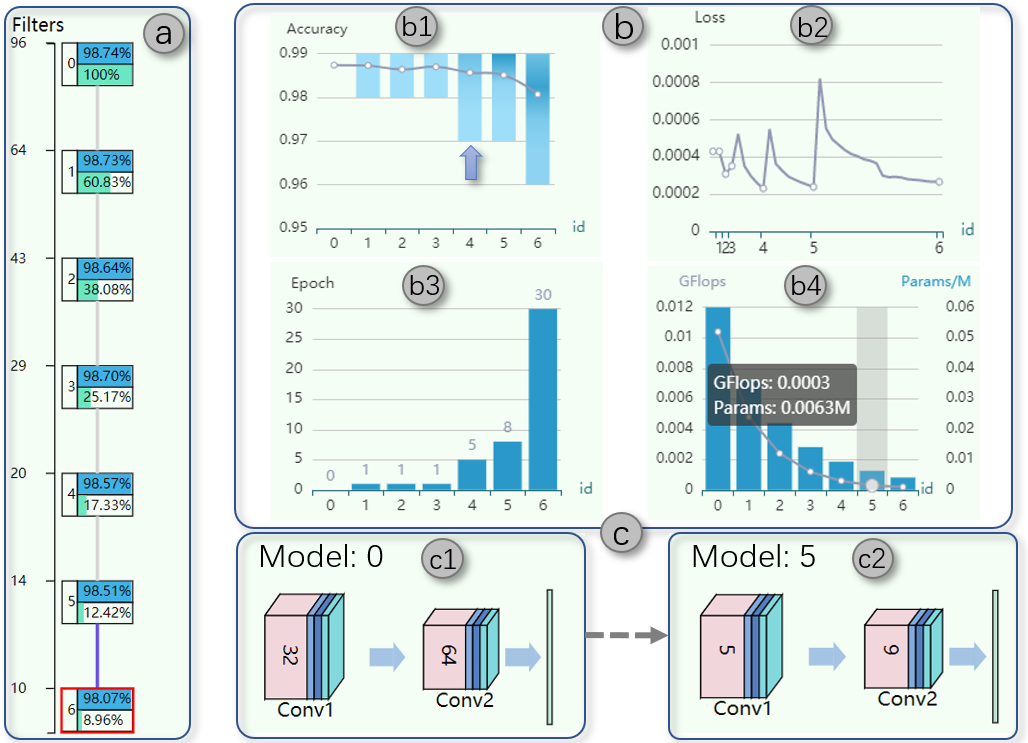}
 \vspace{-0.6cm}
 \caption{The result of CNN pruning. The system executed six prunings to get six models. The \textit{Statistics} view shows the information for Model 6.}
 \label{fig:mnist_result}
\end{figure}

The MNIST dataset \cite{lecun2001gradient} is a commonly used classification dataset. It contains 60,000 images for training and 10,000 images for testing.
We train a two-layer CNN to perform this classification task.
There are 32 filters in the first convolutional layer and 64 filters in the second.
The network structure is shown in \autoref{fig:mnist_result}-c1.
The accuracy of the model is $98.74\%$, and its size is 240KB.

The goal of pruning in this case study is to obtain a pruned CNN with an accuracy of more than $98.50\%$.
The system supports automated pruning and manual pruning.
For this simple CNN, we believe automated pruning is enough to meet the compression goal.

After loading the model to \sysname{}, we need to set some necessary parameters before the pruning. 
First, we configure the dataset parameters to tell the system where the dataset is.
Then, we set the fine-tuning parameters (i.e., set the Delta Loss to $0.000001$, the Target Accuracy to $98.50\%$, the Maximum Epoch to $30$, the Learning Rate to $0.001$, the Optimizer to Adam, and the Batch Size to $100$).
For the setting of the Delta Loss, the Learning Rate, the Optimizer, and the Batch Size, we suggest using the same parameters as the model training stage.
The Delta Loss, the Target Accuracy, and the Maximum Epoch determine the termination of the fine-tuning process.

\begin{figure}
 \centering 
 \includegraphics[width=\columnwidth]{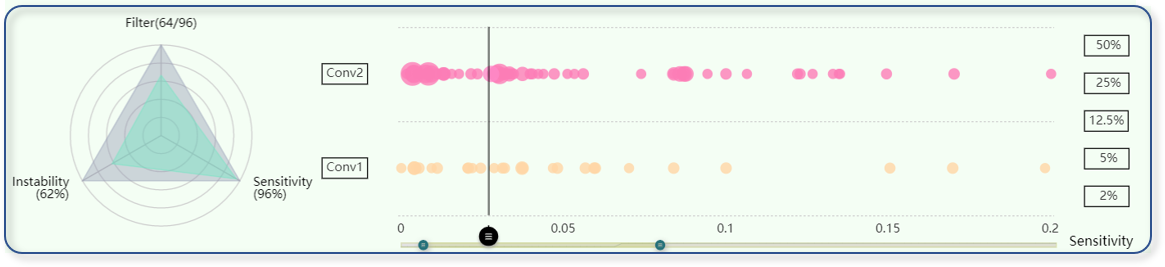}
 \vspace{-0.6cm}
 \caption{The sensitivity and instability distribution of the root model. The radar chart shows the influence of removing one third of the filters.}
 \label{fig:mnist_root_model}
\end{figure}

There is only one node in the \textit{Tree} view after setting the above parameters.
By selecting this root node, we can observe the sensitivity and instability distribution of the model in the \textit{Model} view (Fig. \ref{fig:mnist_root_model}).
For one pruning iteration, we want to minimize the impact of sensitivity while maximally decreasing the instability and the number of filters.
From the estimated pruning results (in the radar chart), we see that removing one-third of the filters will preserve $96\%$ of the sensitivity, and reduce $38\%$ of the instability.
We, therefore, believe we can use the $1/3$ auto-pruning strategy for this case and set the corresponding auto-pruning parameters.
Using the auto-pruning button in the \textit{Tree} view, we automatically prune the model and generate a pruning tree.

\setlength{\belowcaptionskip}{-12pt}
\setlength{\belowcaptionskip}{0pt}

\autoref{fig:mnist_result}-a is the pruning tree for this auto-pruning process. It shows that the number of CNN filters is reduced from 96 to 10 after six pruning iterations.
The prediction accuracy changes marginally in the first five iterations, and the fine-tuning process converges well.
The pruned model from the sixth iteration failed to meet our requirement (i.e., the accuracy dropped to $98.07\% {<} 98.50\%$) and the fine-tuning process did not converge (indicated by the purple color of the line).

\autoref{fig:mnist_result}-b presents the statistics from \sysname{} for further analysis of the auto-pruning process.
\autoref{fig:mnist_result}-b1 shows the recovery ability and the volatility of the six pruned models. As demonstrated by the short and light blue bars, the ``damage'' introduced by the first three pruning operations is small, and the pruned models can easily recover from it.
Starting from the fourth iteration, the resilience of the model decreases, and the accuracy fluctuates more significantly.
\autoref{fig:mnist_result}-b2 shows the model's loss function in the six fine-tuning iterations, which can reduce to the same level after individual fine-tuning iterations. 
For Model 6, pruning has a large impact on the loss, and it cannot recover the accuracy, even after 30 epochs re-training.
Therefore, we think that the parameters of Model 6 are not enough to support the original accuracy.
From the statistics in \autoref{fig:mnist_result}-b1 and \ref{fig:mnist_result}-b2, we believe Model 5 is the best candidate model to meet the compression goal.
\autoref{fig:mnist_result}-b3 and \ref{fig:mnist_result}-b4 show that the number of operations in one forward pass of Model 5 is ${\sim}3$ megaFLOPS, and the number of parameters is ${\sim}6.3$ thousand.
Therefore, the pruning reduced the model's size by $87.58\%$, and the computation cost by $97.12\%$.

The final pruned model (Model 5) is shown in \autoref{fig:mnist_result}-c2, which has five and nine filters in the first and second convolutional layer respectively. 
With {\sysname{}}, we can reveal model-pruning details, such as model convergence, model accuracy, recovery ability, loss fluctuation, and recovery cost. 
These details can help the user better understand the state changes of the model in the pruning process and evaluate the fine-tuning process.
% It is helpful for the user to understand the pruning process, quickly find the target model, and distinguish the anomaly model.

%%%%%%%%%%%%%%%%%%%%%%%%%%%%%%%%%%%%%%%%%%%%%%%%%%%%%%%%%%%%%%%%%%%%%%%%%%%%%%%%%%%%%%%%%%%%%%%%%%%%%
% Progressive interactive pruning
\subsection{Manual Pruning: Flexible Intervention over Pruning}
\begin{figure}
 \centering 
 \includegraphics[width=\columnwidth]{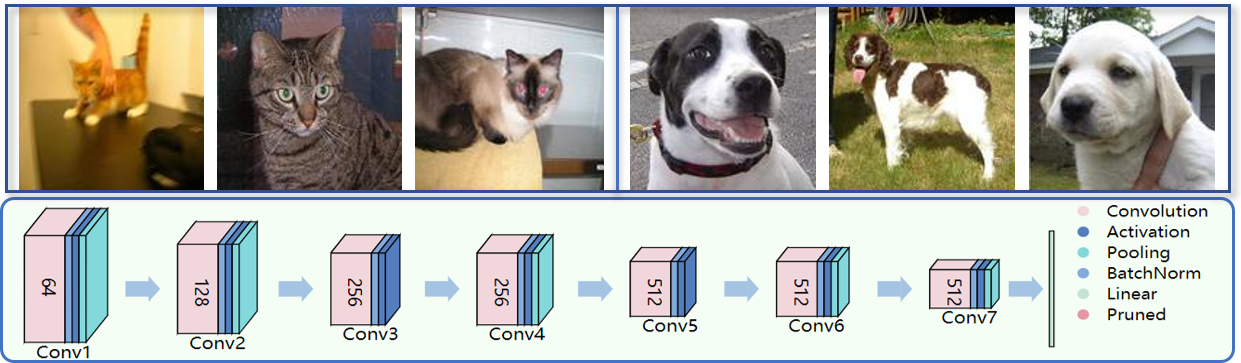}
 \vspace{-0.6cm}
 \caption{The Cat\&Dog dataset and the CNN model architecture.}
 \label{fig:cat_dog_model}
\end{figure}

Our second study presents the case of using the Cat\&Dog dataset \cite{cat_dog_dataset} to interactively achieve a pruning goal. 
The Cat\&Dog dataset contains 25,000 images of cat or dog (the two classes). We randomly select 10,000 cat images and 10,000 dog images as the training dataset. The rest of the images are used for testing.
A CNN with six convolutional layers is trained to differentiate cats from dogs, and its structure is shown in ~\autoref{fig:cat_dog_model}.
The original well-trained model before any compression can achieve a prediction accuracy of 92.76\%.
The model contains 2200 filters, which has 6.88 million parameters with a size of 26.30 MB. A single forward pass of the CNN needs 4.6 GFLOPs operations.

\setlength{\belowcaptionskip}{-12pt}
\setlength{\belowcaptionskip}{0pt}

% The desired pruning goal is to maximally shrink the model while maintaining the prediction accuracy to be above $92.5\%$. To demonstrate the pruning flexibility of \sysname{}, we use manual+estimator pruning in this study, which includes two major stages. The first stage relies on statistical information and immediate visual feedback from the system to remove the filters. The second stage uses the estimator to remove filters in much finer granularity interactively. 

The desired pruning goal is to maximally shrink the model while maintaining the prediction accuracy to be above $92.5\%$. 
\sysname{} can help the user choose the optimal pruning solution by analyzing the pruning process and revealing the pruning details, so as to improve the pruning efficiency and ensures the accuracy of the pruned model. To demonstrate this, we use manual+estimator pruning in this study, which includes two major stages.
The first stage relies on statistical information and immediate visual feedback from the system to remove the filters. The second stage uses the estimator to remove filters interactively in much finer granularity. 
In addition, this section also compares the manual+estimator pruning with the automated only pruning and automated+estimator pruning to show its advantages.

\setlength{\belowcaptionskip}{-15pt}
\begin{figure}[t]
 \centering 
 \includegraphics[width=\columnwidth]{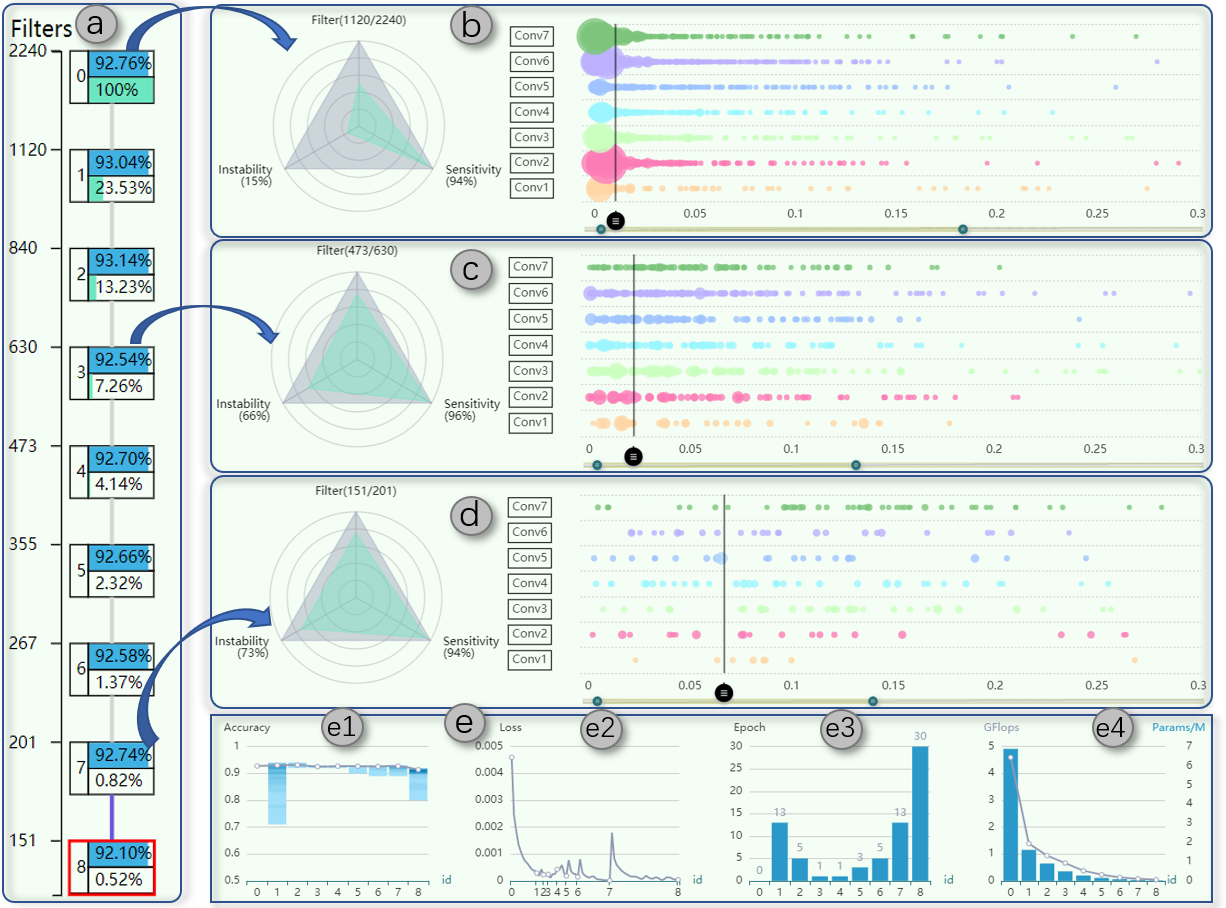}
 \vspace{-0.6cm}
 \caption{The first stage of manual pruning. The \textit{Statistics} view is the information corresponding to Model 8.}
 \label{fig:cat_dog_model_result1}
\end{figure}
\setlength{\belowcaptionskip}{0pt}

\textbf{\textit{Stage 1: Rough-Pruning with Interactive Estimation of Thresholds (R2.2, R2.3).}}
After setting the dataset parameters and fine-tuning parameters, we use the bubble plot in the \textit{Model} view to interactively probe and determine the number of filters to be removed (\autoref{fig:cat_dog_model_result1}).
As shown in \autoref{fig:cat_dog_model_result1}-b, removing 50\% of the filters does not seem to significantly impact the model's sensitivity (change by 6\%) and instability (change by 85\%).
Therefore, we decided to remove 50\% of the filters.
After one round of fine-tuning, we get Model 1 and the statistical information corresponding to this model, as shown in \autoref{fig:cat_dog_model_result1}-e. These statistics reflect the difficulty level of the fine-tuning process.
For example, although the accuracy of Model 1 meets the requirements, the accuracy fluctuated significantly over fine-tuning (reflected by the long strip in \autoref{fig:cat_dog_model_result1}-e1).
Also, the model's training loss reduced a lot over the fine-tuning process (\autoref{fig:cat_dog_model_result1}-e2). With these observations, we decided to remove fewer filters in the next iteration to guarantee a quick recovery. Note that, if the pruned model cannot be recovered after pruning 50\% of the filters, we should restart again from the root node.

In the second pruning iteration, we decided to delete 25\% of the filters (based on our observations of Model 1's statistics).
As expected, the accuracy fluctuation and the training loss changed much less in the pruning from Model 1 to Model 2 (i.e., the second pruning did not damage the model as significantly as the first pruning iteration).

We repeat the above pruning process with on-demand human-interventions until the model no longer meets the requirements. Over this iterative process, we get a pruning tree, as shown in \autoref{fig:cat_dog_model_result1}-a. With the pruning process going forward, the instability of the model gradually decreases (i.e., from \autoref{fig:cat_dog_model_result1}-b, c, to d, the instability changes from 15\%, over 66\% to 73\%). 
Meanwhile, the accuracy fluctuation becomes more and more violent (i.e., from Model 2 to Model 8, the range changes from $92\%{\sim}94\%$ to $80\%{\sim}93\%$), the training loss changes become larger, and the number of required epochs for model recovery increases.
As more and more parameters being removed, the over-parameterization of the model becomes less severe. In addition, using this progressive pruning and evaluation is conducive to manage the model's state change in real-time. Furthermore, users of \sysname{} can directly control the pruning strategy to improve pruning efficiency and prevent the model from being excessively damaged.

\begin{figure}[tb]
 \centering 
 \includegraphics[width=\columnwidth]{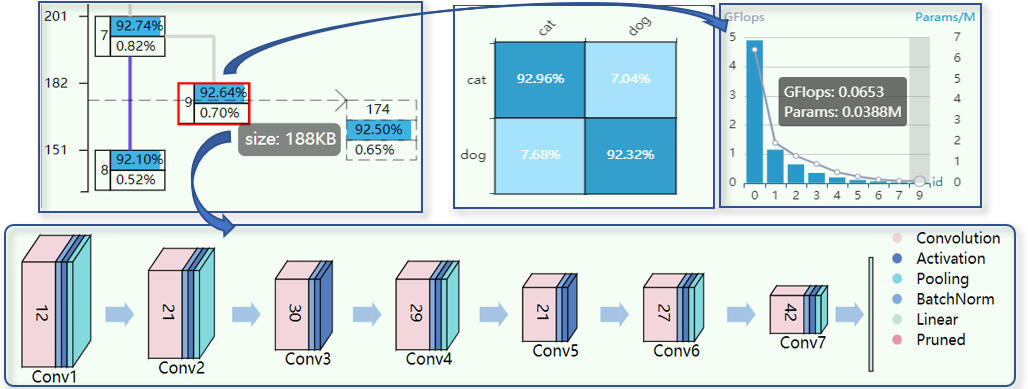}
 \vspace{-0.6cm}
 \caption{The second stage of manual+estimator pruning. The \textit{Statistics} view is the information corresponding to Model 9.} 
 \label{fig:cat_dog_model_result2}
\end{figure}

\textit{\textbf{Stage 2: Fine-Pruning with a Real-Time Estimator (R2.3).}}
From the pruning tree obtained in the first stage (\autoref{fig:cat_dog_model_result1}-a), we can see that the number of filters in the target model should be between that of Model 7 and Model 8.
At this stage, the estimator of \sysname{} can be used to help the user better estimate the number of filters to be removed next.
In the first estimation, the target number of filters given by the estimator is 182.
Therefore, we prune Model 7 to Model 9, i.e., removed 19 (201-182) filters.
Using the estimator again, we find that the number of filters in the target model is 174 (\autoref{fig:cat_dog_model_result2}).
At this time, the gap of filter numbers between the target model and the current model is only 8, so we decided to terminate the pruning.

The pruning process reduced the storage of the model from 26.30 MB to 188 KB.
The accuracy of the final pruned model is 92.64\% (92.96\% for the cat and 92.32\% for the dog, \autoref{fig:cat_dog_model_result2}). 
The accuracy is reduced by 0.12\% compared with the root model.
The parameters of the model are reduced by 99.44\%, and the computation needed for processing an image is reduced by 98.58\%.

\begin{figure}[tb]
 \centering 
 \includegraphics[width=\columnwidth]{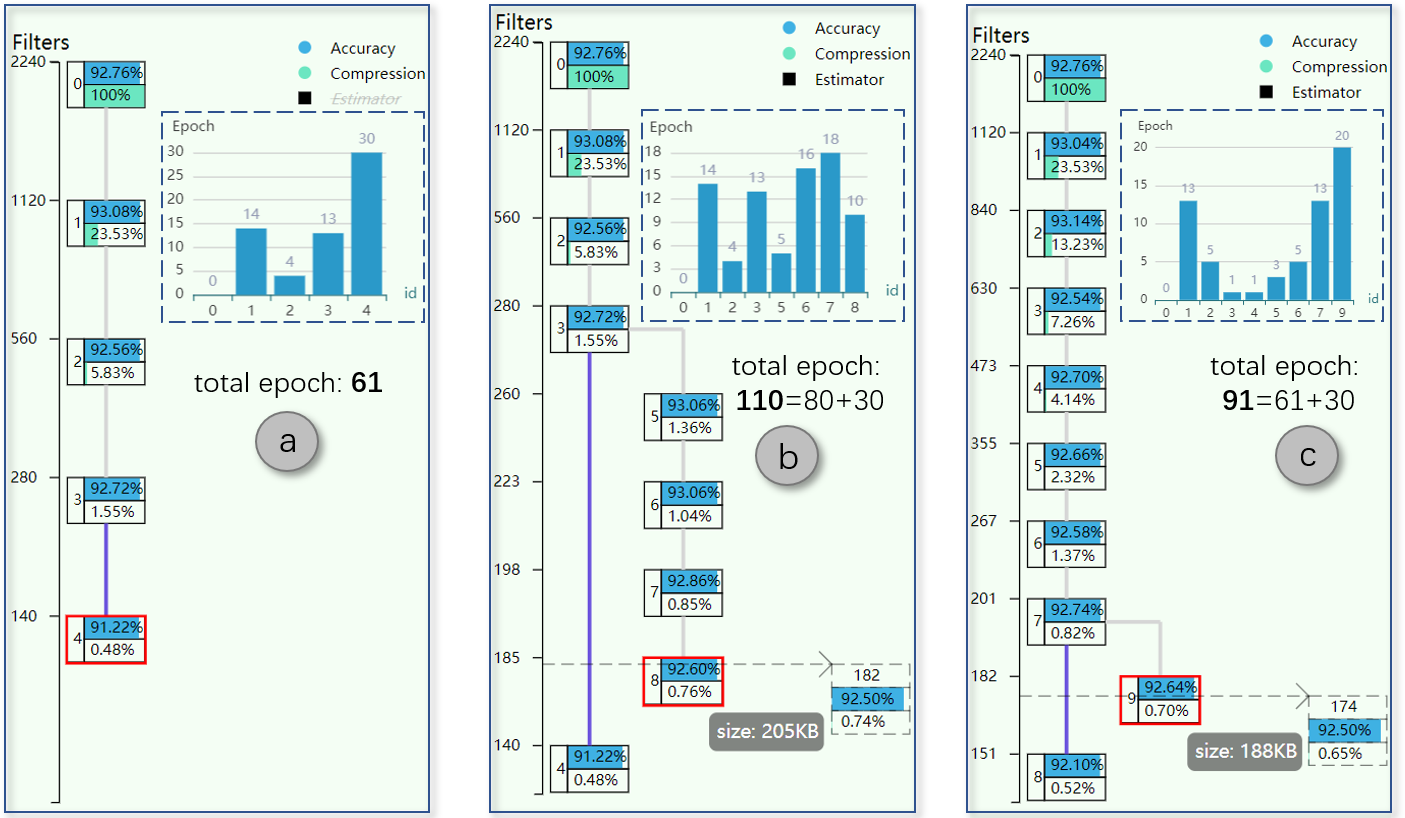}
  \vspace{-0.6cm}
 \caption{
 Comparison of three pruning strategies, (a) automated pruning, (b) automated+estimator pruning, (c) manual+estimator pruning.
 }
 \label{fig:cat_dog_model_result3}
\end{figure}

\textit{\textbf{Comparison of three pruning strategies.}}
To highlight the pruning efficiency of the manual+estimator pruning, we compare it with another two pruning strategies, i.e., the automated pruning and automated+estimator pruning, as shown in \autoref{fig:cat_dog_model_result3}.
The automated pruning in Fig. \ref{fig:cat_dog_model_result3}-a uses the 1/2 auto-pruning plan, i.e., removing half of the filters in each pruning iteration.
From the result, we can see that model 3 is the final pruned model, and the results are worse than the other two strategies. If we use a smaller removal number, e.g., removing 1\% filters in each pruning, we will get a better result, but it will also increase the pruning iterations, costing more computing resources and making the pruning less efficient. Therefore, automated pruning is inflexible and difficult to achieve the best performance.
The automated+estimator pruning in Fig. \ref{fig:cat_dog_model_result3}-b contains two stages. The first stage uses the 1/2 auto-pruning plan and the second stage uses the estimator for finer granularity pruning. 
From the result, we can see that the estimator provides guidance for fine-pruning to help the user get an optimal model. But the large range between Model 3 and Model 4 is not preferable to the second stage of estimation, as it may affect the estimator's performance. Besides the pruning strategy in Fig. \ref{fig:cat_dog_model_result3}-b used about 21\% ($(110-91)/91$, please check the total epoch numbers) of additional time than that of the strategy in Fig. \ref{fig:cat_dog_model_result3}-c (manual+estimator pruning). From these comparisons, we can clearly see how human intervention in the pruning process can help improve the pruning efficiency.

As shown in \cite{frankle2019the}, there should be an optimized sparse sub-network structure in a complex DNN, which can use fewer parameters to get the same accuracy.
Model pruning is an effective way to find this kind of sparse sub-network structure.
Our system targets to detect whether the sub-network has been damaged or not during pruning, and in turn, improve the effectiveness and efficiency of model pruning.

\setlength{\belowcaptionskip}{-6pt}
\setlength{\belowcaptionskip}{0pt}

% Comparison of two pruning strategies. The left one uses the manual pruning, and the right one uses the automated pruning. The total recovery cost of manual pruning is less than that of automated pruning.

%%%%%%%%%%%%%%%%%%%%%%%%%%%%%%%%%%%%%%%%%%%%%%%%%%%%%%%%%%%%%%%%%%%%%%%%%%%%%%%%%%%%%%%%%%%%%%%%%%%%%
\subsection{Diagnose the Pruning Process}

% the case about improving the accuracy of a single category
    % degenerated instances analysis
    % improved instances analysis
    % bad image analysis[wrong label]
    % refine the plan

\begin{figure}[tbh]
 \centering 
 \includegraphics[width=\columnwidth]{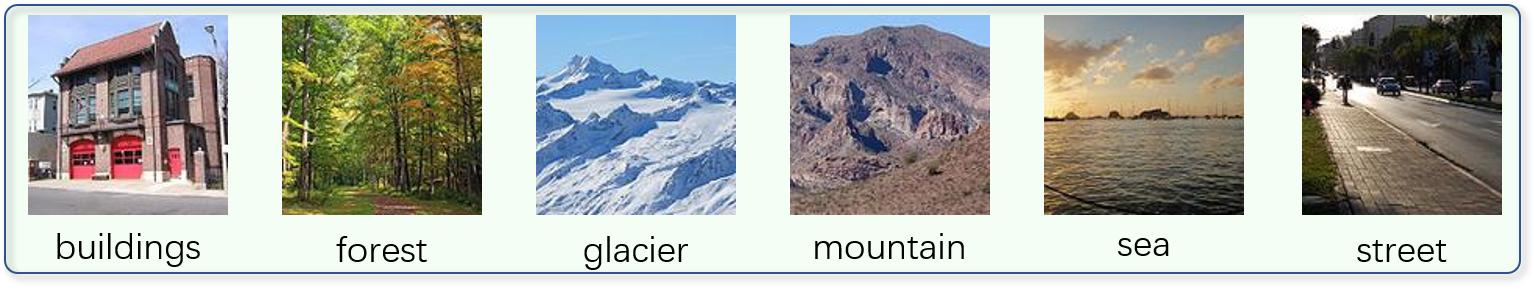}
 \vspace{-0.6cm}
 \caption{Example images from the scene classification dataset.}
 \label{fig:intel_data}
\end{figure}

Our third study presents the case of using an image dataset of nature scenes \cite{scene_classification_dataset} to diagnose the pruning process. 
The dataset contains 17,034 images in 6 classes, 14,034 for training, and 3,000 for testing. 
The 6 categories are: `buildings', `forest', `glacier', `mountain', `sea', and `street'. Example images from individual classes are shown in \autoref{fig:intel_data}.

\setlength{\belowcaptionskip}{-6pt}
\setlength{\belowcaptionskip}{0pt}

\begin{figure}[b]
 \centering 
 \includegraphics[width=\columnwidth]{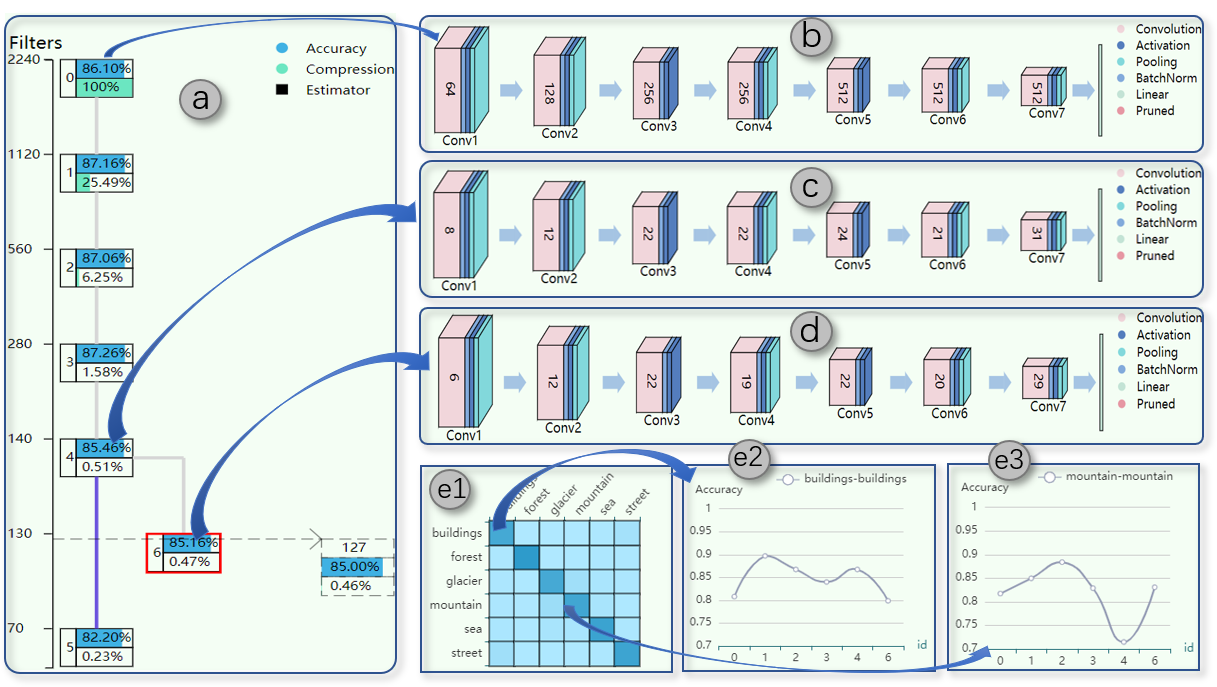}
 \vspace{-0.6cm}
 \caption{The result after model pruning. (e2) The accuracy changes for the class `buildings'. (e3) The accuracy changes for the class `mountain'.}
 \label{fig:intel_result_1}
\end{figure}

A CNN classifier with six convolutional layers is used in this case, and its structure is shown in \autoref{fig:intel_result_1}-b.
The original well-trained model before any compression can achieve a prediction accuracy of 86.10\%.
Our pruning goal is to maximally shrink the model while maintaining the prediction accuracy at above 85.00\%. 
We used \sysname{} to prune the model and got the pruning tree in \autoref{fig:intel_result_1}-a.
After pruning, we reduced the number of filters in the model to 130, and the changes in the model structure are shown in \autoref{fig:intel_result_1}-b,c,d.
Model 6 is our final pruned model, and its accuracy is 85.16\%. 
By analyzing the confusion matrix, we found the model's recognition accuracy for `buildings' dropped sharply from Model 4 to Model 6 (see \autoref{fig:intel_result_1}-e2). 

It is worth mentioning that a model's recognition power for different classes may not be equally important in various tasks. For example, in autonomous driving, recognizing pedestrians around a car is far more important than recognizing the mountains several miles away. Therefore, in some model pruning tasks, domain experts care more about maintaining models' recognition power for certain classes. 
In this case, we use \sysname{} to present an in-depth analysis of the abnormal changes of the accuracy value, and demonstrate how the system can help to refine the pruning plan to reduce its impact.

\setlength{\belowcaptionskip}{-12pt}
\setlength{\belowcaptionskip}{0pt}

\textit{\textbf{Refining Pruning Plan (R3.2).}} \sysname{} can be used to secure the prediction accuracy for the `buildings' class, while maximally compressing the model. From Model 4 to Model 6, the model's overall accuracy descends by 0.3\%,  resulting in 168 degenerated images and 159 improved images. 40 out of the 168 degenerated images and 10 out of the 159 improved images have the true label `buildings'.

\begin{figure}
 \centering 
 \includegraphics[width=\columnwidth]{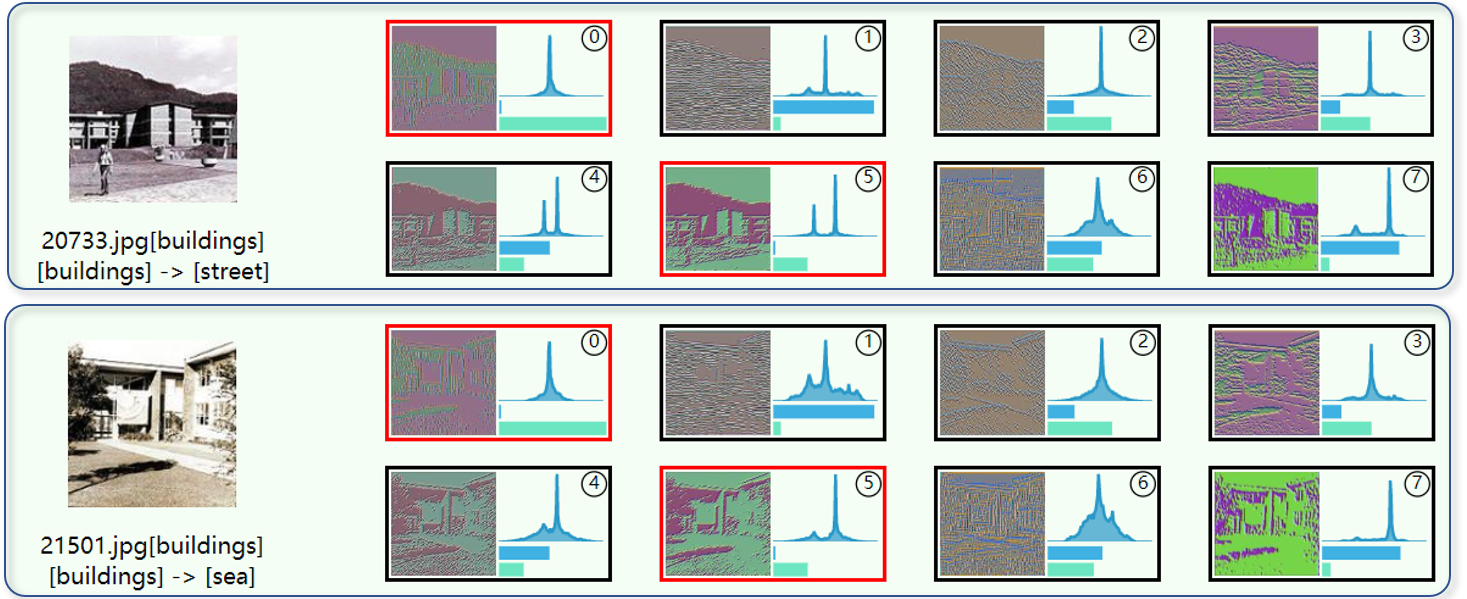}
 %\caption{The examples of degenerated instances in the test dataset from Model 4 to Model 6.}
\vspace{-0.6cm}
 \caption{Examples of degenerated instances from Model 4 to Model 6.}
 \label{fig:intel_degeneration_result}
\end{figure}

We analyze the degenerated `buildings' instances to find out why pruning affects the recognition of this particular class.
\autoref{fig:intel_degeneration_result} shows two  degenerated instances of the class `buildings'.
From the filter visualization matrix, we can see that the system deletes the filters that have the lowest sensitivity and highest instability, i.e., Filter 0 and Filter 5 (see the blue and green bar on the right of the filter visualization).
However, to the class of `buildings', the features captured by these two filters are not the least important.
The area chart in the upper right of the filter visualization displays the distribution of pixel values for the filter visualization image (feature map).
In general, the more concentrated the distribution is, the sharper the features are extracted.
Comparing the eight distributions, Filter 1 and Filter 6 are the least important ones (for `buildings'). 
The pixel value distributions for these two filters are more chaotic than others, and there are more noises in the corresponding feature maps. The decision of deleting Filter 0 and 5, rather than Filter 1 and 6, reduces the model's power in recognizing `buildings', which is hard to recover from the subsequent fine-tuning process.

% Our goal is to help deep learning experts to refine their pruning plan from Model 4 to Model 6 without significantly sacrificing the model's accuracy for class `buildings'.
Based on the above observation, we decided to refine the pruning plan by removing Filter 1 and 6, but keep Filter 0 and 5.
We set up a new branch from Model 4 and pruned it with the refined plan to get Model 7, and the result is shown in \autoref{fig:teaser}.
The accuracy of Model 7 is 85.40\% and 87.41\% for the class `buildings'.
Therefore, our system optimized the pruning plan through this in-depth analysis of the filters.

% discussion, verified by multi-experiments
To avoid the influence of randomness introduced during the fine-turning process, we repeated the pruning multiple times to validate if our refined pruning plan is indeed better. 
Specifically, we pruned Model 4 20 times, 10 times of which used the original plan, the other 10 times used the refined plan.
After pruning, we got 20 pruned models. Their statistics are shown in \autoref{tab:statistical_two_plan}.
From the table, we can see that the refined pruning plan can effectively reduce the decreasing trend of the accuracy of class `buildings' when pruning Model 4.

\begin{table}[]
\begin{tabular}{|c|c|c|c|}
\hline
\multicolumn{2}{|c|}{}                          & Original Plan & Refined Plan \\ \hline
\multirow{3}{*}{All Categories}     & Degenerated & 158.3         & 157.0        \\ \cline{2-4} 
                                 & Improved    & 149.7         & 152.1        \\ \cline{2-4} 
                                 & Accuracy     & 85.18\%       & 85.30\%      \\ \hline
\multirow{3}{*}{`buildings' Only} & Degenerated & 35.1          & 20.2         \\ \cline{2-4} 
                                 & Improved    & 12.1          & 17.8         \\ \cline{2-4} 
                                 & Accuracy     & \textbf{81.46\%}       & \textbf{86.17\%}      \\ \hline
\end{tabular}
\caption{The statistics of two pruning plans (averaged over 10 runs).}
\label{tab:statistical_two_plan}
\vspace{-0.6cm}
\end{table}

\begin{figure}[tbh]
 \centering 
 \includegraphics[width=\columnwidth]{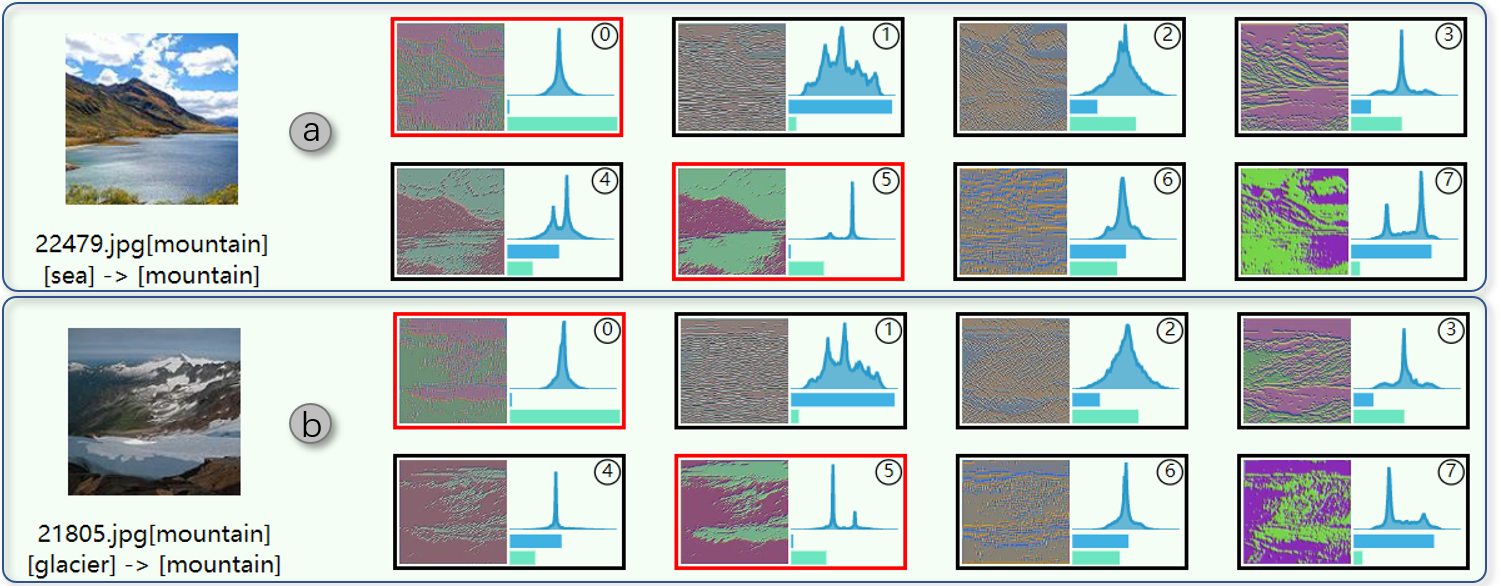}
 %\caption{The examples of improved instances in the test dataset from Model 4 to Model 6.}
 \vspace{-0.6cm}
 \caption{Examples of the improved instances from Model 4 to Model 6.}
 \label{fig:intel_evolution}
\end{figure}

% About 50%(40 images) of the evolution instances changed from glaciers to mountains.
\textit{\textbf{Interpreting Pruning Process (R3.1).}} From Model 4 to Model 6, the accuracy for the class `mountain' increased by 11.62\% (\autoref{fig:intel_result_1}-e3), resulting in 20 degenerated instances and 81 improved instances for this class. With \sysname{}, we can interpret what has contributed to the model improvement over the pruning.
As shown in \autoref{fig:intel_evolution}, we selected some images to analyze why the pruning plan improved the accuracy of `mountain'.
The image in \autoref{fig:intel_evolution}-a was mis-classified as `sea' by the model initially. The pruning removed Filter 5, which extracted the majority of the pixels for `sea' in the image. As a result, the pruned model believes the image is more like a `mountain', rather than `sea'. 
Similarly, in \autoref{fig:intel_evolution}-b, Filter 5 mostly extracted the glacier features, which is probably why the image was mis-classified as `glacier' before pruning. Removing these noisy features makes the model concentrate more on the mountain and generate the correct prediction of `mountain'.

\begin{figure}[tb]
 \centering 
 \includegraphics[width=\columnwidth]{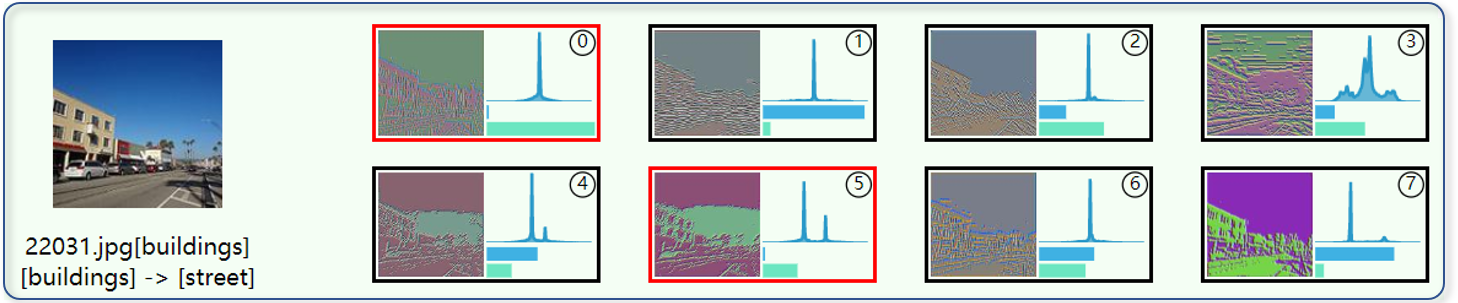}
 %\caption{The confusing image example in the test dataset from Model 4 to Model 6.}
 \vspace{-0.6cm}
 \caption{The confusing image example from Model 4 to Model 6.}
 \label{fig:intel_degeneration_bad_image}
\end{figure}

\textit{\textbf{Identify Confusing Images.}} Additionally, from the investigations with the degenerated image instances (from Model 4 to Model 6) with \sysname{}, we also found images with improper labels. 
For example, the image in \autoref{fig:intel_degeneration_bad_image} is one of the degenerated instances with the true label `buildings'.
The original image contains both street and buildings, and the street takes a major portion of the image. Although the image is labeled as `buildings', we feel `street' is more proper for it.
As this image only confuses the model, we recommend removing it from the test dataset, which can make our model evaluation more objective.

\setlength{\belowcaptionskip}{-12pt}
\setlength{\belowcaptionskip}{0pt}

\section{Discussion and Domain Experts' Feedback}
% {1) expert feedback.}

%{2). discuss strengths and weaknesses.}

We conducted open-ended interviews with two machine learning experts ($E_1$, $E_2$) to discuss the strengths, weaknesses, and potential extensions of \sysname{}. The experts' research interests are accelerating deep neural networks, and model pruning is an important portion of their research works.
We first introduce the design goal of \sysname{} and the individual visualization components to them (in about 30 minutes).
With the experts' background on model pruning, they can quickly pick up the domain-related concepts and understand the functionality of individual components, though it still took them some time to get familiar with the visualization and interaction of the system (about 60 minutes).
We then go through the cases presented in the Case Studies and ask them to freely play with the system and provide feedback.
% based on their domain experience with model pruning.

In general, both experts felt positive on \sysname{}, and they believed that the model pruning process can be clearly and intuitively presented through visualization techniques.
$E_1$ likes the \textit{Tree} view the most, as it can reveal the evolution of the pruned model quickly and allow users to reprocess the pruning interactively. The estimator in the \textit{Tree} view was very interesting to him, and he agreed that it could effectively help users determine the pruning depth in the last pruning stage.
$E_2$ appreciated the progressive pruning method proposed in \sysname{}. Through the proposed criteria (i.e., recovery capability, loss fluctuation, and recovery cost), domain experts can evaluate the pruning process more objectively.
Both $E_1$ and $E_2$ were glad to see the effectiveness of the \textit{Filter} view in interpreting CNNs and refining pruning plans. With the existing techniques, it is still hard for them to thoroughly understand the model pruning process from numerical statistics only. \sysname{} provides a practical way for them to interpret individual filters visually and understand their roles over the pruning process.
Moreover, both experts agreed that the concepts of degenerated and improved instances are beneficial in effectively identifying images of interest.

The experts also pointed out several limitations of \sysname{}, as well as some improvements that can be applied in the future. For example, $E_1$ mentioned that for models with many classes, the \textit{Confusion Matrix} view may not scale well. We plan to improve this view by supporting the filtering of different classes in the future.
Also, the experts provided their domain feedback on how we can proceed further along this research direction.
$E_1$ suggested that we can extend model pruning to fully connected layers, as the parameters from these layers can take a considerably large portion of the networks in many scenarios.
$E_2$ recommended us to enhance the system by supporting the comparisons of different pruning criteria. As model pruning is still a fast-growing topic, he believed more and more criteria will be proposed. With our system, researchers can more intuitively compare different pruning plans, which in turn, will help them optimize the pruning process.

\section{Conclusion}
In this work, we proposed {\sysname}, a visual analytics system to help machine learning experts to understand, diagnose, and refine the CNN pruning process.
{\sysname} contains four visualization components that work together to reveal model details on different levels over the iterative pruning process.
Two criteria and three metrics are used in \sysname{} to estimate filters' importance before pruning and evaluate the pruned model's quality after pruning. Both the pre-estimation and post-evaluation facilitate users to make and refine their pruning plans. Moreover, the capability of \sysname{} in thoroughly examining the degenerated and improved data instances within one pruning iteration plays an essential role in interpreting and diagnosing the pruned model. Through multiple case studies on CNN models with real-world sizes, we validated the effectiveness of \sysname{}.

\acknowledgments{
%The authors wish to thank A, B, and C. This work was supported in part by a grant from XYZ (\# sss-sss).
This work was supported by the Strategic Priority Research Program of the Chinese Academy of Sciences, grant No. XDA19080102. The work was started at The Ohio State University when Guan was visiting the GRAVITY research group. The authors would like to thank all GRAVITY members for their suggestions and insightful discussions.
% , and we would thank all people in the lab who participated in the discussion.
}

\bibliographystyle{abbrv}

\bibliography{template}

\begin{thebibliography}{10}

\bibitem{cat_dog_dataset}
{Dataset: Dogs vs. Cats} (accessed: 2020-03-10).
\newblock \url{https://www.kaggle.com/c/dogs-vs-cats/data}.

\bibitem{scene_classification_dataset}
{Dataset: scene classification dataset} (accessed: 2020-03-10).
\newblock \url{https://www.kaggle.com/puneet6060/intel-image-classification}.

\bibitem{flask_web}
{Flask} (accessed: 2020-03-10).
\newblock \url{https://palletsprojects.com/p/flask}.

\bibitem{PyTorch_framework}
{PyTorch} (accessed: 2020-03-10).
\newblock \url{https://pytorch.org}.

\bibitem{bilal2018do}
A.~{Bilal}, A.~{Jourabloo}, M.~{Ye}, X.~{Liu}, and L.~{Ren}.
\newblock Do convolutional neural networks learn class hierarchy.
\newblock {\em IEEE Transactions on Visualization and Computer Graphics},
  24(1):152--162, 2017.

\bibitem{carreira-perpinan2018learning}
M.~A. {Carreira-Perpinan} and Y.~{Idelbayev}.
\newblock ``learning-compression'' algorithms for neural net pruning.
\newblock In {\em 2018 IEEE/CVF Conference on Computer Vision and Pattern
  Recognition}, pages 8532--8541, 2018.

\bibitem{choo2018visual}
J.~{Choo} and S.~{Liu}.
\newblock Visual analytics for explainable deep learning.
\newblock {\em IEEE Computer Graphics and Applications}, 38(4):84--92, 2018.

\bibitem{dong2017learning}
X.~{Dong}, S.~{Chen}, and S.~J. {Pan}.
\newblock Learning to prune deep neural networks via layer-wise optimal brain
  surgeon.
\newblock In {\em Advances in Neural Information Processing Systems}, pages
  4857--4867, 2017.

\bibitem{dubey2018coreset}
A.~{Dubey}, M.~{Chatterjee}, and N.~{Ahuja}.
\newblock Coreset-based neural network compression.
\newblock In {\em Proceedings of the European Conference on Computer Vision
  (ECCV)}, pages 469--486, 2018.

\bibitem{frankle2019the}
J.~{Frankle} and M.~{Carbin}.
\newblock The lottery ticket hypothesis: Finding sparse, trainable neural
  networks.
\newblock In {\em ICLR 2019 : 7th International Conference on Learning
  Representations}, 2019.

\bibitem{girshick2014rich}
R.~{Girshick}, J.~{Donahue}, T.~{Darrell}, and J.~{Malik}.
\newblock Rich feature hierarchies for accurate object detection and semantic
  segmentation.
\newblock In {\em CVPR '14 Proceedings of the 2014 IEEE Conference on Computer
  Vision and Pattern Recognition}, pages 580--587, 2014.

\bibitem{goodfellow2014generative}
I.~{Goodfellow}, J.~{Pouget-Abadie}, M.~{Mirza}, B.~{Xu}, D.~{Warde-Farley},
  S.~{Ozair}, A.~{Courville}, and Y.~{Bengio}.
\newblock Generative adversarial nets.
\newblock In {\em Advances in Neural Information Processing Systems 27}, pages
  2672--2680, 2014.

\bibitem{guo2016dynamic}
Y.~{Guo}, A.~{Yao}, and Y.~{Chen}.
\newblock Dynamic network surgery for efficient dnns.
\newblock In {\em NIPS'16 Proceedings of the 30th International Conference on
  Neural Information Processing Systems}, pages 1387--1395, 2016.

\bibitem{han2016deep}
S.~{Han}, H.~{Mao}, and W.~J. {Dally}.
\newblock Deep compression: Compressing deep neural networks with pruning,
  trained quantization and huffman coding.
\newblock In {\em ICLR 2016 : International Conference on Learning
  Representations 2016}, 2016.

\bibitem{han2015learning}
S.~{Han}, J.~{Pool}, J.~{Tran}, and W.~J. {Dally}.
\newblock Learning both weights and connections for efficient neural networks.
\newblock In {\em NIPS'15 Proceedings of the 28th International Conference on
  Neural Information Processing Systems}, pages 1135--1143, 2015.

\bibitem{he2018soft}
Y.~{He}, G.~{Kang}, X.~{Dong}, Y.~{Fu}, and Y.~{Yang}.
\newblock Soft filter pruning for accelerating deep convolutional neural
  networks.
\newblock In {\em IJCAI 2018: 27th International Joint Conference on Artificial
  Intelligence}, pages 2234--2240, 2018.

\bibitem{he2019filter}
Y.~{He}, P.~{Liu}, Z.~{Wang}, Z.~{Hu}, and Y.~{Yang}.
\newblock Filter pruning via geometric median for deep convolutional neural
  networks acceleration.
\newblock In {\em 2019 IEEE/CVF Conference on Computer Vision and Pattern
  Recognition (CVPR)}, pages 4340--4349, 2019.

\bibitem{he2017channel}
Y.~{He}, X.~{Zhang}, and J.~{Sun}.
\newblock Channel pruning for accelerating very deep neural networks.
\newblock In {\em 2017 IEEE International Conference on Computer Vision
  (ICCV)}, pages 1398--1406, 2017.

\bibitem{kahng2019gan}
M.~{Kahng}, N.~{Thorat}, D.~H.~P. {Chau}, F.~B. {Viegas}, and M.~{Wattenberg}.
\newblock Gan lab: Understanding complex deep generative models using
  interactive visual experimentation.
\newblock {\em IEEE Transactions on Visualization and Computer Graphics},
  25(1):310--320, 2018.

\bibitem{krizhevsky2017imagenet}
A.~{Krizhevsky}, I.~{Sutskever}, and G.~E. {Hinton}.
\newblock Imagenet classification with deep convolutional neural networks.
\newblock {\em Communications of The ACM}, 60(6):84--90, 2017.

\bibitem{lecun2015deep}
Y.~{LeCun}, Y.~{Bengio}, and G.~{Hinton}.
\newblock Deep learning.
\newblock {\em Nature}, 521(7553):436--444, 2015.

\bibitem{lecun2001gradient}
Y.~{Lecun}, L.~{Bottou}, Y.~{Bengio}, and P.~{Haffner}.
\newblock Gradient-based learning applied to document recognition.
\newblock {\em Intelligent Signal Processing}, pages 306--351, 2001.

\bibitem{lecun1990optimal}
Y.~LeCun, J.~S. Denker, and S.~A. Solla.
\newblock Optimal brain damage.
\newblock In {\em Advances in neural information processing systems}, pages
  598--605, 1990.

\bibitem{li2017pruning}
H.~{Li}, A.~{Kadav}, I.~{Durdanovic}, H.~{Samet}, and H.~P. {Graf}.
\newblock Pruning filters for efficient convnets.
\newblock In {\em ICLR 2017 : International Conference on Learning
  Representations 2017}, 2017.

\bibitem{liu2018analyzing}
M.~{Liu}, J.~{Shi}, K.~{Cao}, J.~{Zhu}, and S.~{Liu}.
\newblock Analyzing the training processes of deep generative models.
\newblock {\em IEEE Transactions on Visualization and Computer Graphics},
  24(1):77--87, 2018.

\bibitem{liu2017towards}
M.~{Liu}, J.~{Shi}, Z.~{Li}, C.~{Li}, J.~{Zhu}, and S.~{Liu}.
\newblock Towards better analysis of deep convolutional neural networks.
\newblock {\em IEEE Transactions on Visualization and Computer Graphics},
  23(1):91--100, 2017.

\bibitem{liu2017learning}
Z.~{Liu}, J.~{Li}, Z.~{Shen}, G.~{Huang}, S.~{Yan}, and C.~{Zhang}.
\newblock Learning efficient convolutional networks through network slimming.
\newblock In {\em 2017 IEEE International Conference on Computer Vision
  (ICCV)}, pages 2755--2763, 2017.

\bibitem{luo2017thinet}
J.-H. {Luo}, J.~{Wu}, and W.~{Lin}.
\newblock Thinet: A filter level pruning method for deep neural network
  compression.
\newblock In {\em 2017 IEEE International Conference on Computer Vision
  (ICCV)}, pages 5068--5076, 2017.

\bibitem{ming2019rulematrix}
Y.~{Ming}, H.~{Qu}, and E.~{Bertini}.
\newblock Rulematrix: Visualizing and understanding classifiers with rules.
\newblock {\em IEEE Transactions on Visualization and Computer Graphics},
  25(1):342--352, 2019.

\bibitem{mnih2015human}
V.~{Mnih}, K.~{Kavukcuoglu}, D.~{Silver}, A.~A. {Rusu}, J.~{Veness}, M.~G.
  {Bellemare}, A.~{Graves}, M.~{Riedmiller}, A.~K. {Fidjeland}, G.~{Ostrovski},
  S.~{Petersen}, C.~{Beattie}, A.~{Sadik}, I.~{Antonoglou}, H.~{King},
  D.~{Kumaran}, D.~{Wierstra}, S.~{Legg}, and D.~{Hassabis}.
\newblock Human-level control through deep reinforcement learning.
\newblock {\em Nature}, 518(7540):529--533, 2015.

\bibitem{molchanov2017pruning}
P.~{Molchanov}, S.~{Tyree}, T.~{Karras}, T.~{Aila}, and J.~{Kautz}.
\newblock Pruning convolutional neural networks for resource efficient
  inference.
\newblock In {\em ICLR 2017 : International Conference on Learning
  Representations 2017}, 2017.

\bibitem{pezzotti2018deepeyes}
N.~{Pezzotti}, T.~{Hollt}, J.~V. {Gemert}, B.~P. {Lelieveldt}, E.~{Eisemann},
  and A.~{Vilanova}.
\newblock Deepeyes: Progressive visual analytics for designing deep neural
  networks.
\newblock {\em IEEE Transactions on Visualization and Computer Graphics},
  24(1):98--108, 2018.

\bibitem{ren2017squares}
D.~{Ren}, S.~{Amershi}, B.~{Lee}, J.~{Suh}, and J.~D. {Williams}.
\newblock Squares: Supporting interactive performance analysis for multiclass
  classifiers.
\newblock {\em IEEE Transactions on Visualization and Computer Graphics},
  23(1):61--70, 2017.

\bibitem{seide2011conversational}
F.~{Seide}, G.~{Li}, and D.~{Yu}.
\newblock Conversational speech transcription using context-dependent deep
  neural networks.
\newblock In {\em INTERSPEECH}, pages 437--440, 2011.

\bibitem{simonyan2015very}
K.~{Simonyan} and A.~{Zisserman}.
\newblock Very deep convolutional networks for large-scale image recognition.
\newblock In {\em ICLR 2015 : International Conference on Learning
  Representations 2015}, 2015.

\bibitem{springenberg2014striving}
J.~T. {Springenberg}, A.~{Dosovitskiy}, T.~{Brox}, and M.~A. {Riedmiller}.
\newblock Striving for simplicity: The all convolutional net.
\newblock In {\em ICLR (workshop track)}, 2014.

\bibitem{tai2016convolutional}
C.~{Tai}, T.~{Xiao}, Y.~{Zhang}, X.~{Wang}, and W.~{E}.
\newblock Convolutional neural networks with low-rank regularization.
\newblock In {\em ICLR 2016 : International Conference on Learning
  Representations 2016}, 2016.

\bibitem{tung2018clip}
F.~{Tung} and G.~{Mori}.
\newblock Clip-q: Deep network compression learning by in-parallel
  pruning-quantization.
\newblock In {\em 2018 IEEE/CVF Conference on Computer Vision and Pattern
  Recognition}, pages 7873--7882, 2018.

\bibitem{maaten2008visualizing}
L.~van~der {Maaten} and G.~{Hinton}.
\newblock Visualizing data using t-sne.
\newblock {\em Journal of Machine Learning Research}, 9:2579--2605, 2008.

\bibitem{wang2018dqnviz}
J.~Wang, L.~Gou, H.-W. Shen, and H.~Yang.
\newblock Dqnviz: A visual analytics approach to understand deep q-networks.
\newblock {\em IEEE transactions on visualization and computer graphics},
  25(1):288--298, 2018.

\bibitem{wang2018ganviz}
J.~{Wang}, L.~{Gou}, H.~{Yang}, and H.-W. {Shen}.
\newblock Ganviz : A visual analytics approach to understand the adversarial
  game.
\newblock {\em IEEE Transactions on Visualization and Computer Graphics},
  24(6):1905--1917, 2018.

\bibitem{wang2019deepvid}
J.~{Wang}, L.~{Gou}, W.~{Zhang}, H.~{Yang}, and H.-W. {Shen}.
\newblock Deepvid : Deep visual interpretation and diagnosis for image
  classifiers via knowledge distillation.
\newblock {\em IEEE Transactions on Visualization and Computer Graphics},
  25(6):2168--2180, 2019.

\bibitem{yu2018nisp}
R.~{Yu}, A.~{Li}, C.-F. {Chen}, J.-H. {Lai}, V.~I. {Morariu}, X.~{Han},
  M.~{Gao}, C.-Y. {Lin}, and L.~S. {Davis}.
\newblock Nisp: Pruning networks using neuron importance score propagation.
\newblock In {\em 2018 IEEE/CVF Conference on Computer Vision and Pattern
  Recognition}, pages 9194--9203, 2018.

\bibitem{yuan2021survey}
J.~Yuan, C.~Chen, W.~Yang, M.~Liu, J.~Xia, and S.~Liu.
\newblock A survey of visual analytics techniques for machine learning.
\newblock {\em Computational Visual Media}, 7(1), 2021.

\bibitem{zhang2019manifold}
J.~{Zhang}, Y.~{Wang}, P.~{Molino}, L.~{Li}, and D.~S. {Ebert}.
\newblock Manifold: A model-agnostic framework for interpretation and diagnosis
  of machine learning models.
\newblock {\em IEEE Transactions on Visualization and Computer Graphics},
  25(1):364--373, 2019.

\bibitem{zhang2016accelerating}
X.~{Zhang}, J.~{Zou}, K.~{He}, and J.~{Sun}.
\newblock Accelerating very deep convolutional networks for classification and
  detection.
\newblock {\em IEEE Transactions on Pattern Analysis and Machine Intelligence},
  38(10):1943--1955, 2016.

\end{thebibliography}

\end{document}